\documentclass[letterpaper]{article}
\usepackage{aaai}
\usepackage{times}
\usepackage{helvet}
\usepackage{courier}
\usepackage{amssymb}
\usepackage{multirow}
\usepackage{bm}
\usepackage[boxed]{algorithm}
\usepackage[noend]{algorithmic}
\usepackage[pdftex]{epsfig}
\newtheorem{theorem}{Theorem}
\newtheorem{lemma}[theorem]{Lemma}
\newtheorem{corollary}[theorem]{Corollary}
\newtheorem{definition}[theorem]{Definition}

\begin{document}
\title{Information-Theoretic Approach to Efficient Adaptive Path Planning for\\ Mobile Robotic Environmental Sensing}
\author{Kian Hsiang Low$^{\dag}$\and John M. Dolan$^{\dag\S}$\and Pradeep Khosla$^{\dag\S}$\\
Department of Electrical and Computer Engineering$^{\dag}$, Robotics Institute$^{\S}$\\
Carnegie Mellon University\\ 
5000 Forbes Avenue, Pittsburgh, PA 15213, USA\\ 
\{bryanlow, jmd\}@cs.cmu.edu, pkk@ece.cmu.edu
}

\newcommand{\ma}[1]{{\bf #1}}
\newcommand{\ve}[1]{{\mathbf #1}}
\newcommand{\set}[1]{{\mathcal #1}}
% The ``defined as'' symbol.
\newcommand{\defeq}[0]{\ensuremath{\stackrel{\triangle}{=}}}

\newcommand{\squishlist}{
 \begin{list}{$\bullet$}
  { \setlength{\itemsep}{-2pt}
     \setlength{\parsep}{3pt}
     \setlength{\topsep}{3pt}
     \setlength{\partopsep}{0pt}
     \setlength{\leftmargin}{0.7em}
     \setlength{\labelwidth}{0.5em}
     \setlength{\labelsep}{0.2em} } }

\newcommand{\squishlisttwo}{
 \begin{list}{$\bullet$}
  { \setlength{\itemsep}{0pt}
     \setlength{\parsep}{0pt}
    \setlength{\topsep}{0pt}
    \setlength{\partopsep}{0pt}
    \setlength{\leftmargin}{2em}
    \setlength{\labelwidth}{1.5em}
    \setlength{\labelsep}{0.5em} } }

\newcommand{\squishend}{
  \end{list}  }

\maketitle
\begin{abstract}
\vspace{-0.5mm}
\begin{quote}
Recent research in robot exploration and mapping has focused on sampling environmental hotspot fields.
This exploration task is formalized by \citeauthor{LowAAMAS08} \shortcite{LowAAMAS08} in a sequential decision-theoretic planning under uncertainty framework called MASP. The time complexity of solving MASP approximately depends on the map resolution, which limits its use in large-scale, high-resolution exploration and mapping.
To alleviate this computational difficulty,
this paper presents an information-theoretic approach to MASP ($i$MASP) for efficient adaptive path planning; by reformulating the cost-minimizing $i$MASP as a reward-maximizing problem, its time complexity becomes independent of map resolution and is less sensitive to increasing robot team size as demonstrated both theoretically and empirically.
Using the reward-maximizing dual, we derive a novel adaptive variant of maximum entropy sampling, thus improving the induced exploration policy performance.
It also allows us to establish theoretical bounds quantifying the performance advantage of optimal adaptive over non-adaptive policies and the performance quality of approximately optimal vs. optimal adaptive policies. 
We show analytically and empirically the superior performance of $i$MASP-based policies for sampling the log-Gaussian process to that of policies for the widely-used Gaussian process in mapping the hotspot field.
Lastly, we provide sufficient conditions that, when met, guarantee adaptivity has no benefit under an assumed environment model.
\vspace{-0.5mm}
\end{quote}
\end{abstract}

\section{Introduction}
%\vspace{-1mm}
Recent research in multi-robot exploration and mapping \cite{LowAAMAS08,Guestrin07} has focused on sampling environmental fields, some of which typically feature a few small \emph{hotspots} in a large region \cite{Webster01}. Such a \emph{hotspot field} 
often arises in environmental and ecological sensing applications such as precision agriculture, mineral prospecting, monitoring of ocean phenomena, forest ecosystems, pollution, or contamination. In particular, the hotspot field (e.g., plankton density and mineral distribution in Fig.~\ref{fig:pkmap}) is characterized by \emph{continuous, positively skewed, spatially correlated} measurements with the hotspots exhibiting extreme measurements and much higher spatial variability than the rest of the field. With limited (e.g., point-based) robot sensing range, a complete coverage becomes impractical in terms of resource costs (e.g., energy consumption). 
So, to accurately map the field, the hotspots have to be sampled at a higher resolution.

The hotspot field discourages static sensor placement \cite{Guestrin05} because a large number of sensors has to be positioned to detect and refine the sampling of hotspots. If these static sensors are not placed in any hotspot initially, they cannot reposition by themselves to locate one. In contrast, a robot team is capable of performing high-resolution hotspot sampling due to its mobility.
Hence, it is desirable to build a mobile robot team that can actively explore to map a hotspot field. 

To learn a hotspot field map, the \emph{exploration strategy} of the robot team has to plan resource-constrained observation paths that minimize the map uncertainty of the hotspot field.
To achieve this, the recent work of \citeauthor{LowAAMAS08} \shortcite{LowAAMAS08} has proposed such a strategy that plans non-myopic adaptive paths to minimize the uncertainty of a spatial model of the hotspot field. In particular, both (a) modeling and (b) planning components are designed to fully exploit the environmental structure in order to yield a high-quality map:
(a) The hotspot field is assumed to be realized from a non-parametric probabilistic model called the log-Gaussian process, which can provide a formal measure of map uncertainty and more importantly, characterize the abovementioned hotspot field measurements well; (b) The exploration task is formalized in a sequential decision-theoretic planning under uncertainty framework, which we call the \emph{\underline{m}ulti-robot \underline{a}daptive \underline{s}ampling \underline{p}roblem} (MASP).
So, MASP can be viewed as a sequential, non-myopic version of active learning.
In contrast to finite-state Markov decision problems, MASP adopts a more complex but realistic continuous-state, \emph{non-Markovian} problem structure so that its induced exploration policy can be informed by the complete history of continuous, spatially correlated observations for selecting paths.
It is unique in unifying formulations of exploration problems along the entire adaptivity (see Def.~\ref{def:2}) spectrum, 
thus subsuming existing non-adaptive formulations and allowing the performance advantage of a more adaptive policy to be theoretically realized.
Through MASP, it is demonstrated that a more adaptive strategy can exploit clustering phenomena in a hotspot field to produce lower map uncertainty.

However, MASP is besieged by a serious computational drawback due to its measure of map uncertainty using the mean-squared error criterion. Consequently, the time complexity of solving MASP (approximately) depends on the map resolution, which limits its practical use in large-scale, high-resolution exploration and mapping.

The principal contribution of this paper is to alleviate this computational difficulty through an information-theoretic approach to MASP ($i$MASP) for efficient adaptive path planning, which measures map uncertainty based on the entropy criterion instead.
Unlike MASP, reformulating the cost-minimizing $i$MASP as a reward-maximizing problem causes its time complexity of being solved approximately to be independent of the map resolution and less sensitive to larger robot team size as demonstrated both theoretically and empirically in this paper. 
Additional contributions stemming from this reward-maximizing formulation include:
\squishlist
\item transforming the commonly-used non-adaptive maximum entropy sampling problem \cite{Shewry87} into a novel adaptive variant, thus improving the performance of the induced exploration policy;
%elaborate on entropy criterion not new, but the adaptive variant is new.
\item establishing theoretical bounds to quantify the performance advantage of optimal adaptive over non-adaptive exploration policies, and the performance quality of approximately optimal vs. optimal adaptive policies;
\item given an assumed environment model (e.g., occupancy grid map), establishing sufficient conditions that, when met, guarantee adaptivity provides no benefit; and
\item showing analytically and empirically the superior performance of $i$MASP-based policies for sampling the log-Gaussian process ($\ell$GP) to that of policies for the widely-used Gaussian process (GP) \cite{Guestrin05,Shewry87,Guestrin07} in mapping the hotspot field.\vspace{1mm}
\squishend
%\subsection*{Related Work}
\noindent
{\bf Related Work.}
Beyond its computational gain, $i$MASP retains the beneficial properties of MASP: it is novel in the class of model-based exploration strategies to perform both wide-area coverage and hotspot sampling. The former considers sparsely sampled areas to be of high uncertainty and thus spreads the observations evenly across the environmental field. The latter expects areas of high uncertainty to contain highly-varying measurements and hence produces clustered observations. 
Like MASP, $i$MASP also covers the entire adaptivity spectrum, thus subsuming the existing non-adaptive entropy-based problem formulation \cite{Shewry87}.
In contrast, all other model-based strategies \cite{Guestrin07b,Guestrin07} are non-adaptive and achieve only wide-area coverage;
they are observed to perform well only with smoothly-varying fields.
Similar to MASP, $i$MASP can plan non-myopic multi-robot paths, which are more desirable than greedy or single-robot paths \cite{Guestrin07b,Guestrin07}.
%In contrast to model-based strategies that select observations to reduce map uncertainty, observation selection in design-based strategies \cite{LowICRA07,Nowak06} is constrained by the sampling design, which is not devised to consider resource costs. As a result, the locations have to be chosen by the strategy first before planning to minimize the resource costs to sample them.
%This entails ``sunk'' costs in motion: the observation paths have to traverse terrain that does not require sampling to reach the selected locations.
%These strategies often require multiple ``passes'' through the region of interest such that new locations are adaptively selected and sampled in each pass.
% Hence, they are partially adaptive (Definition~\ref{def:2}).
%In our exploration task, the cost of motion is assumed to outweigh that of sampling/sensing, thus making these strategies extremely cost-inefficient.
%Modifying a design-based strategy to involve resource costs may invalidate the estimators associated with the strategy.
\vspace{-1mm}
\section{Cost-Minimizing Problem Formulations}%\vspace{-1mm}
\label{sect:pf}
We formalize here the information-theoretic exploration problems at the two extremes of the adaptivity spectrum. Exploration problems residing within the spectrum can be formalized in a similar manner.
Note that the use of the entropy criterion in non-myopic active learning is not new but is limited to the non-adaptive problem formulation \cite{Shewry87}, which is presented here as a comparison to the novel adaptive problem formulation.
It can be observed that the resulting cost-minimizing formulations differ from that of MASP by only the entropy criterion.
However, as we shall see in a later section, their reward-maximizing dual formulations are significantly different from that of MASP in terms of interpretation and computational complexity.\vspace{1mm}\\
{\bf Notation and Preliminaries}.
Let $\set{X}$ be the domain of the hotspot field corresponding to a finite set of grid cell locations. 
An observation taken (e.g., by a single robot) at stage $i$ comprises a pair of location $x_i \in \set{X}$ and its measurement $z_{x_i}$.
More generally, $k$ observations taken (e.g., by $k$ robots or 1 robot taking $k$ observations) at stage $i$ can be represented by a pair of vectors $\ve{x}_i$ of $k$ locations and $\ve{z}_{\ve{x}_i}$ of the corresponding measurements.
% Let $Z_{x_i}$ and $\ve{Z}_{\ve{x}_i}$ be the random measurements of the respective realizations $z_{x_i}$ and $\ve{z}_{\ve{x}_i}$.
\vspace{0mm}
\begin{definition}[Posterior Data]
The posterior data $d_i$ at stage $i>0$ comprises\vspace{-1mm}
\squishlist
\item the prior data $d_0 = \langle\ve{x}_0, \ve{z}_{\ve{x}_0}\rangle$ available at stage $0$, and
\item a complete history of observations $\ve{x}_1, \ve{z}_{\ve{x}_1}, \ldots, \ve{x}_i, \ve{z}_{\ve{x}_i}$ induced by $k$ observations per stage over stages $1$ to $i$.\vspace{0mm}
\squishend
\end{definition}
Let $\ve{x}_{0:i}$ and $\ve{z}_{\ve{x}_{0:i}}$ denote vectors comprising the location and measurement components of the posterior data $d_i$ (i.e., concatenations of $\ve{x}_0, \ve{x}_1, \ldots, \ve{x}_i$ and $\ve{z}_{\ve{x}_0}, \ve{z}_{\ve{x}_1}, \ldots, \ve{z}_{\ve{x}_i}$), respectively.
Let $\overline{\ve{x}}_{0:i}$ denote the vector comprising locations of domain $\set{X}$ not observed in $d_i$, and $\ve{z}_{\overline{\ve{x}}_{0:i}}$  be the vector comprising the corresponding measurements.
Let $Z_{x_i}$, $\ve{Z}_{\ve{x}_i}$, $\ve{Z}_{\ve{x}_{0:i}}$, $\ve{Z}_{\overline{\ve{x}}_{0:i}}$ be the random measurements corresponding to the respective realizations $z_{x_i}$, $\ve{z}_{\ve{x}_i}$, $\ve{z}_{\ve{x}_{0:i}}$, $\ve{z}_{\overline{\ve{x}}_{0:i}}$.
\begin{definition}[Characterizing Adaptivity]
Suppose prior data $d_0$ are available and $n$ new locations are to be explored. Then, an exploration strategy is\vspace{-1mm}
\squishlist
\item {\bf adaptive} if its policy to select each vector $\ve{x}_{i+1}$ of $k$ new locations  depends only on the previously sampled data $d_{i}$ for $i = 0,\ldots,n/k-1$. So, this strategy selects k observations per stage over $n/k$ stages.
If $k = 1$, this strategy is strictly adaptive. Increasing $k$ makes it partially adaptive;
\item {\bf non-adaptive} if its policy to select each new location $x_{i+1}$ for $i = 0,\ldots,n-1$ is independent of the measurements $z_{x_1},\ldots,z_{x_n}$. As a result, all $n$ new locations $x_{1},\ldots,x_{n}$ can be selected prior to exploration. That is, this strategy selects all $n$ observations in a single stage.\vspace{-1mm}
\squishend
\label{def:2}
\end{definition}
{\bf Objective Function}.
The exploration objective is to plan observation paths that minimize the uncertainty of mapping the hotspot field.
To achieve this, we use the entropy criterion to measure map uncertainty.
Given the posterior data $d_n$, the \emph{posterior map entropy} of domain $\set{X}$ can be represented by the posterior joint entropy of the measurements $\ve{Z}_{\overline{\ve{x}}_{0:n}}$ at the unobserved locations $\overline{\ve{x}}_{0:n}$:\vspace{-1mm}
\begin{equation}
\hspace{0mm}\mathbb{H}[\ve{Z}_{\overline{\ve{x}}_{0:n}}|d_n] \hspace{-0.1mm}\defeq\hspace{0mm} - \hspace{-1mm}\displaystyle \int\hspace{-1mm} f(\ve{z}_{\overline{\ve{x}}_{0:n}}|d_n) \log f(\ve{z}_{\overline{\ve{x}}_{0:n}}|d_n) \ d \ve{z}_{\overline{\ve{x}}_{0:n}}\vspace{-1mm}
\label{eq:1a}
\end{equation} 
where $f$ denotes a probability density function.\vspace{1mm}\\
{\bf Value Function}.
If only the prior data $d_0$ are available, an exploration strategy has to produce a policy for 
selecting observation paths that minimize the \emph{expected} posterior map entropy instead. 
This policy must then collect the optimal observations $\ve{x}_{1}, \ve{z}_{\ve{x}_{1}}, \ldots, \ve{x}_{n}, \ve{z}_{\ve{x}_{n}}$ during exploration to form posterior data $d_n$.
The value under an exploration policy $\pi$ is defined to be the expected posterior map entropy (i.e., expectation of (\ref{eq:1a})) when starting in $d_0$ and following $\pi$ thereafter:\vspace{-2mm}
\begin{equation}
\begin{array}{rl}
\hspace{-0mm}V^{\pi}_{0}(d_0) \defeq& \displaystyle\mathbb{E}\{ \mathbb{H}[\ve{Z}_{\overline{\ve{x}}_{0:n}}|d_n] | d_0, \pi\}\\
=& \hspace{-0mm}\displaystyle \int \hspace{-0mm}f(\ve{z}_{\ve{x}_{1:n}}|d_0,\pi) \ \mathbb{H}[\ve{Z}_{\overline{\ve{x}}_{0:n}}|d_n] \ \mbox{d}\ve{z}_{\ve{x}_{1:n}} \ .\vspace{-1mm}
\end{array}
\label{eq:1b}
\end{equation}
The strategies of \citeauthor{Guestrin05} \shortcite{Guestrin05} and \citeauthor{Guestrin07} \shortcite{Guestrin07} have optimized a closely related \emph{mutual information} criterion that measures the expected entropy reduction of unobserved locations $\overline{\ve{x}}_{0:n}$ by observing $\ve{x}_{1:n}$ (i.e., $\mathbb{H}[\ve{Z}_{\overline{\ve{x}}_{0:n}}|d_0] - \mathbb{E}\{ \mathbb{H}[\ve{Z}_{\overline{\ve{x}}_{0:n}}|d_n]|d_0\}$).
This is deficient for the exploration objective because mutual information may be maximized by a choice of $\ve{x}_{1:n}$ inducing a very large prior entropy $\mathbb{H}[\ve{Z}_{\overline{\ve{x}}_{0:n}}|d_0]$ but not necessarily the smallest expected posterior map entropy $\mathbb{E}\{ \mathbb{H}[\ve{Z}_{\overline{\ve{x}}_{0:n}}| d_n]|d_0\}$.

In the next two subsections, we will describe how the adaptive and non-adaptive exploration policies can be derived to minimize the expected posterior map entropy (\ref{eq:1b}).\vspace{1mm}\\
{\bf Adaptive Exploration}. 
The adaptive policy $\pi$ for directing a team of $k$ robots is structured to collect $k$ observations per stage over a finite planning horizon of $n$ stages. This implies each robot observes $1$ location per stage and is thus constrained to explore at most $n$ new locations over $n$ stages.
Formally, 
$\pi \defeq\langle\pi_0(d_0),\ldots, \pi_{n-1}(d_{n-1})\rangle$ where $\pi_i:d_i$ $\rightarrow\ve{a}_i$ maps data $d_i$ to a vector of robots' actions $\ve{a}_i \in \set{A}(\ve{x}_i)$ at stage $i$,
and $\set{A}(\ve{x}_i)$ is the joint action space of the robots given their current locations $\ve{x}_i$. We assume the transition function $\tau : \ve{x}_i \times \ve{a}_i \rightarrow \ve{x}_{i+1}$ \emph{deterministically} moves the robots to their next locations $\ve{x}_{i+1}$ at stage $i+1$.
Combining $\pi_i$ and $\tau$ gives $\ve{x}_{i+1} \leftarrow \tau(\ve{x}_{i}, \pi_i(d_i))$.
%, thus implying $\pi$ is adaptive (Def.~\ref{def:2}).
We can observe from this assignment that the sequential (i.e., stagewise) selection of $k$ new locations $\ve{x}_{i+1}$ to be included in the observation paths depends only on the previously sampled data $d_i$ along the paths for stage $i = 0,\ldots,n-1$.
Hence, policy $\pi$ is adaptive (Def.~\ref{def:2}).

Solving the adaptive exploration problem $i$MASP($1$) means choosing the adaptive policy $\pi$ to minimize $V^{\pi}_{0}(d_0)$ (\ref{eq:1b}), which we call the \emph{optimal adaptive policy} $\pi^{1}$.
That is, 
%$\pi^{1}$ is induced by the \emph{optimal value function}
$V^{\pi^{1}}_{0}(d_0) = \min_{\pi} V^{\pi}_{0}(d_0)$.
Plugging $\pi^{1}$ into (\ref{eq:1b}) gives the $n$-stage dynamic programming equations:\vspace{-2mm}
%\vspace{-3mm}
\begin{equation}
\hspace{-1.5mm}
\begin{array}{rl}
V^{\pi^1}_{i}(d_i) =& \hspace{-3mm}\displaystyle \int f(\ve{z}_{\ve{x}_{i+1}}| d_i,\pi^1_i) \ V^{\pi^1}_{i+1}(d_{i+1}) \ \mbox{d}\ve{z}_{\ve{x}_{i+1}} \\
=& \hspace{-3mm}\displaystyle \int f(\ve{z}_{\tau(\ve{x}_{i}, \pi^1_i(d_i))}| d_i) \ V^{\pi^1}_{i+1}(d_{i+1}) \ \mbox{d}\ve{z}_{\tau(\ve{x}_{i}, \pi^1_i(d_i))} \\
=& \hspace{-3mm}\displaystyle \min_{\ve{a}_{i} \in \set{A}(\ve{x}_{i})} \int \hspace{-1mm}f(\ve{z}_{\tau(\ve{x}_{i}, \ve{a}_{i})}| d_i) \ V^{\pi^1}_{i+1}(d_{i+1}) \ \mbox{d}\ve{z}_{\tau(\ve{x}_{i}, \ve{a}_{i})} \\
V^{\pi^1}_{n}(d_{n})
=& \hspace{-1mm}\displaystyle \mathbb{H}[\ve{Z}_{\overline{\ve{x}}_{0:n}}| d_n]\vspace{-3.5mm}
%\vspace{-2mm}
\end{array}
\label{eq:4}
\end{equation}
for stage $i = 0, \ldots, n-1$. 
The first and second equalities follow from $f(\ve{z}_{\ve{x}_{1:n}}|d_0,\pi^1) = \Pi^{n-1}_{i=0} f(\ve{z}_{\ve{x}_{i+1}}| d_i,\pi^1_i)$ and $\ve{x}_{i+1} \leftarrow \tau(\ve{x}_{i}, \pi^1_i(d_i))$ respectively. 
%Policy $\pi^{1}$ can be obtained from the optimal actions under each stagewise minimum.\vspace{1mm}\\
Policy $\pi^{1} = \langle\pi^{1}_0(d_0),\ldots, \pi^{1}_{n-1}(d_{n-1})\rangle$ can be determined in a stagewise manner by\vspace{-3.5mm}
%\vspace{-4mm}
%\begin{equation}
$$\pi^{1}_{i}(d_{i}) = \mathop{\arg\min}_{\ve{a}_{i} \in \set{A}(\ve{x}_{i})} \int f(\ve{z}_{\tau(\ve{x}_{i}, \ve{a}_{i})} | d_i) \ V^{\pi^1}_{i+1}(d_{i+1}) \ \mbox{d}\ve{z}_{\tau(\ve{x}_{i}, \ve{a}_{i})} \ .\vspace{-2mm}$$
% \ .
%\label{eq:4oap}
%\end{equation}
%From (\ref{eq:4oap}), 
Note that the optimal action $\pi^{1}_0(d_0)$ at stage $0$ can be determined prior to exploration using prior data $d_0$. However, each action rule $\pi^{1}_i(d_{i})$ at stage $i = 1, \ldots, n-1$ defines the optimal action to take in response to $d_{i}$, part of which (i.e., $\ve{x}_{1}, \ve{z}_{\ve{x}_{1}}, \ldots, \ve{x}_{i}, \ve{z}_{\ve{x}_{i}}$) are only observed during exploration.\vspace{1mm}\\
{\bf Non-Adaptive Exploration}. 
The non-adaptive policy $\pi$ is structured to collect, in $1$ stage, $n$ observations per robot with a team of $k$ robots.
So, each robot is also constrained to explore at most $n$ new locations, but they have to do this within $1$ stage. Formally, $\pi \defeq \pi_0(d_0)$ where $\pi_0 : d_0 \rightarrow \ve{a}_{0:n-1}$ maps prior data $d_0$ to a vector $\ve{a}_{0:n-1}$
of action components concatenating a sequence of robots' actions $\ve{a}_0, \ldots, \ve{a}_{n-1}$.
Combining $\pi_0$ and $\tau$ gives $\ve{x}_{1:n} \leftarrow \tau(\ve{x}_{0:n-1}, \pi_0(d_0))$.
%, thus implying $\pi$ is non-adaptive (Def.~\ref{def:2}).
We can observe from this assignment that the selection of $k\times n$ new locations $\ve{x}_1, \ldots, \ve{x}_{n}$ to form the observation paths are independent of the measurements $\ve{z}_{\ve{x}_1}, \ldots, \ve{z}_{\ve{x}_n}$ obtained along the paths during exploration.
Hence, policy $\pi$ is non-adaptive (Def.~\ref{def:2}) and all new locations can be selected in a single stage prior to exploration.

Solving the non-adaptive exploration problem $i$MASP($n$) involves choosing $\pi$ to minimize $V^{\pi}_{0}(d_0)$ (\ref{eq:1b}), which we call the \emph{optimal non-adaptive policy} $\pi^{n}$
(i.e., 
%$\pi^{n}$ is induced by the optimal value function
$V^{\pi^{n}}_{0}(d_0) = \min_{\pi} V^{\pi}_{0}(d_0)$).
Plugging $\pi^{n}$ into (\ref{eq:1b}) gives the 1-stage equation:
\vspace{-2.5mm}
\begin{equation}
\begin{array}{l}
\hspace{-2mm}V^{\pi^n}_{0}(d_0) = \displaystyle \int f(\ve{z}_{\ve{x}_{1:n}}| d_0,\pi^n_0) \  \mathbb{H}[\ve{Z}_{\overline{\ve{x}}_{0:n}}| d_n] \ \mbox{d}\ve{z}_{\ve{x}_{1:n}} \\
\hspace{-2mm}= \hspace{-1mm}\displaystyle \int f(\ve{z}_{\tau(\ve{x}_{0:n-1}, \pi^n_{0}(d_0))}| d_0) \  \mathbb{H}[\ve{Z}_{\overline{\ve{x}}_{0:n}}| d_n] \ \mbox{d}\ve{z}_{\tau(\ve{x}_{0:n-1}, \pi^n_{0}(d_0))} \\
\hspace{-2mm}= \hspace{-1mm}\displaystyle \min_{\ve{a}_{0:n-1}} \hspace{-1mm}\int \hspace{-1mm}f(\ve{z}_{\tau(\ve{x}_{0:n-1}, \ve{a}_{0:n-1})}| d_0) \  \mathbb{H}[\ve{Z}_{\overline{\ve{x}}_{0:n}}| d_n] \ \mbox{d}\ve{z}_{\tau(\ve{x}_{0:n-1}, \ve{a}_{0:n-1})} .\vspace{-2mm}
\end{array}
\label{eq:3}
\end{equation}
The second equality follows from $\ve{x}_{1:n} \leftarrow \tau(\ve{x}_{0:n-1}, \pi^n_{0}(d_0))$ described above.
%Policy $\pi^{n}$ can be determined from the optimal action sequence under the minimum prior to exploration.\vspace{-1mm}
Policy $\pi^{n} = \pi^{n}_0(d_0)$ can therefore be determined in a single stage by $\pi^{n}_{0}(d_{0}) =$ 
\vspace{-1mm}
%\begin{equation}
$$\mathop{\arg\min}_{\ve{a}_{0:n-1}}\hspace{-1mm} \int\hspace{-1mm} f(\ve{z}_{\tau(\ve{x}_{0:n-1}, \ve{a}_{0:n-1})}|d_0) \  \mathbb{H}[\ve{Z}_{\overline{\ve{x}}_{0:n}}| d_n] \ \mbox{d}\ve{z}_{\tau(\ve{x}_{0:n-1}, \ve{a}_{0:n-1})}\vspace{-1mm} \ .$$
%\vspace{-1mm}
%\label{eq:3oap}
%\end{equation}
Note that the optimal sequence of robots' actions $\pi^{n}_{0}(d_{0})$ (i.e., optimal observation paths) can be determined prior to exploration since the prior data $d_0$ are available.\vspace{-1.5mm}
\section{Reward-Maximizing Dual Formulations}
%\vspace{-1mm}
\label{sect:afMASP}
%\vspace{-4mm}
In this section, we transform the cost-minimizing $i$MASP($1$) (\ref{eq:4}) and $i$MASP($n$) (\ref{eq:3}) into reward-maximizing problems and show their equivalence.
The reward-maximizing $i$MASP($n$) turns out to be the well-known \emph{maximum entropy sampling} (MES) problem \cite{Shewry87}: \vspace{-2mm}
\begin{equation}
%\begin{array}{rl}
 U^{\pi^n}_{0}(d_{0}) = \max_{\ve{a}_{0:n-1}} \mathbb{H}[\ve{Z}_{\tau(\ve{x}_{0:n-1}, \ve{a}_{0:n-1})}| d_{0}] \ ,\vspace{-2mm}
%\end{array}
\label{eq:3r}
\end{equation}
which is a single-staged problem of selecting $k\times n$ new locations $\ve{x}_1, \ldots, \ve{x}_n$ with maximum entropy to form the observation paths.
This dual ensues from the equivalence result $V^{\pi^n}_0(d_0) = \mathbb{H}[\ve{Z}_{\overline{\ve{x}}_{0}}| d_0] - U^{\pi^n}_0(d_0)$ relating cost-minimizing and reward-maximizing $i$MASP($n$)'s in the non-adaptive exploration setting,
%\vspace{-3mm}
which follows from the chain rule of entropy. This result says the original objective of minimizing expected posterior map entropy (i.e., $V^{\pi^n}_0(d_0)$ (\ref{eq:3})) is equivalent to that of discharging from prior map entropy $\mathbb{H}[\ve{Z}_{\overline{\ve{x}}_{0}}| d_0]$ the largest entropy into the selected paths (i.e., $U^{\pi^n}_0(d_0)$ (\ref{eq:3r})).
Hence, their optimal non-adaptive policies coincide.
% \cite{Cover91}

Our reward-maximizing $i$MASP($1$) is a novel adaptive variant of MES. Unlike the cost-minimizing $i$MASP($1$), it can be subject to convex analysis, which allows monotone-bounding approximations to be developed as shown later.
It comprises the following $n$-stage dynamic programming equations:\vspace{-3mm}
\begin{equation}
\hspace{-2mm}
\begin{array}{rl}
\displaystyle U^{\pi^1}_{i}(d_{i}) =& \hspace{-1mm}\displaystyle\max_{\ve{a}_{i} \in \set{A}(\ve{x}_{i})}
\mathbb{H}[\ve{Z}_{\tau(\ve{x}_{i}, \ve{a}_{i})}| d_{i}] \ + \\
& \hspace{-1mm}\displaystyle\int f(\ve{z}_{\tau(\ve{x}_{i}, \ve{a}_{i})}| d_i) \ U^{\pi^1}_{i+1}(d_{i+1}) \ \mbox{d}\ve{z}_{\tau(\ve{x}_{i}, \ve{a}_{i})}\\
\displaystyle U^{\pi^1}_{t}(d_{t}) =& \hspace{-1mm}\displaystyle\max_{\ve{a}_{t} \in \set{A}(\ve{x}_t)}
\mathbb{H}[\ve{Z}_{\tau(\ve{x}_{t}, \ve{a}_{t})}| d_{t}]\vspace{-3mm}
\end{array}
\label{eq:4r}
\end{equation}
for stage $i = 0, \ldots, t-1$ where $t = n - 1$. Each stagewise reward reflects the entropy of $k$ new locations $\ve{x}_{i+1}$ to be potentially selected into the paths.
By maximizing the sum of expected rewards over $n$ stages in (\ref{eq:4r}),
the reward-maximizing $i$MASP($1$) absorbs the largest expected entropy into the selected paths.
In the adaptive exploration setting, the cost-minimizing and reward-maximizing $i$MASP($1$)'s are also equivalent (i.e., their optimal adaptive policies coincide):\vspace{-2mm}
\begin{theorem} $V^{\pi^1}_i(d_{i}) =
\displaystyle\mathbb{H}[\ve{Z}_{\overline{\ve{x}}_{0:i}}| d_i] - U^{\pi^1}_{i}(d_{i})$
for stage $i = 0, \ldots, n-1$.\vspace{-1mm}
\label{thm:1}
\end{theorem}
The work of \citeauthor{LowAAMAS08} \shortcite{LowAAMAS08} has also provided an equivalence result to relate the cost-minimizing and reward-maximizing MASPs through the use of the variance decomposition formula in its induction proof.
In contrast, the induction proof to Theorem~\ref{thm:1} uses the chain rule of entropy,
which entails a computational complexity reduction (not available to MASP) as described next.

%
%Its interpretation is similar to that of non-adaptive exploration except that it holds over all $n$ stages rather than just for stage $0$.
%The reward-maximizing problem structure can also be exploited by the approximation algorithms to gain computational advantage of being independent of the map resolution as detailed in the rest of this paper.
In cost-minimizing $i$MASP($1$),
% and $i$MASP($n$), 
the time complexity of evaluating the cost (i.e., posterior map entropy (\ref{eq:1a})) depends on the domain size $|\set{X}|$ for the environment models described in the next section. By transforming into the dual, the time complexity of evaluating each stagewise reward becomes independent of $|\set{X}|$ because it reflects only the uncertainty of the new locations to be potentially selected into the observation paths.
As a result, the runtime of the approximation algorithm proposed in a later section does not depend on the map resolution, which is clearly advantageous in large-scale, high-resolution exploration and mapping.
In contrast, the reward-maximizing MASP \cite{LowAAMAS08} utilizing the mean-squared error criterion does not share this computational advantage, as the time needed to evaluate each stagewise reward still depends on $|\set{X}|$.
We will evaluate this computational advantage using time complexity analysis in a later section.\vspace{-2mm}
\section{Learning the Hotspot Field Map}
%\vspace{-1.5mm}
\label{sect:model}
%\vspace{-3mm}
Traditionally, a hotspot is defined as a location where its measurement exceeds a pre-defined extreme. 
But, hotspot locations do not usually occur in isolation but in clusters.
So, it is useful to characterize hotspots with spatial properties.
Accordingly, we define a hotspot field to vary as a realization of a spatial random field $\{Y_x>0\}_{x \in \set{X}}$ such that putting together the observed measurements of a realization $\{y_x\}_{x \in \set{X}}$
gives a positively skewed 1D sample frequency distribution (e.g., Fig.~\ref{fig:planktonsim}b).
In this section, we will highlight the problem with modeling the hotspot field directly using GP and explain how the $\ell$GP remedies this.
We will also show analytically that the $i$MASP-based policy for sampling $\ell$GP is adaptive and exploits clustering phenomena but that for sampling GP lacks these properties.
\vspace{1mm}\\
% 
%
%\noindent
{\bf Gaussian Process}.
A widely-used random field to model environmental phenomena is the GP \cite{Guestrin05,Guestrin07b,Guestrin07}.
The stationary assumption on the GP covariance structure is very sensitive to strong positive skewness of hotspot field measurements (e.g., Fig.~\ref{fig:planktonsim}b) and is easily violated by a few extreme ones \cite{Webster01}.
In practice, this can cause reconstructed fields to display large hotspots centered about a few extreme observations and prediction variances to be unrealistically small in hotspots, which are undesirable.
%Hence, if the GP is used to directly model a hotspot field exhibiting positively skewed measurements (e.g., Fig.~\ref{fig:plankton}), the assumption of stationary covariance structure will most likely be violated. 
So, if GP is used to model a hotspot field directly, it may not map well.
To remedy this, a standard statistical practice 
is to take the log of the measurements (i.e., $Z_x=\log Y_x$) to remove skewness and extremity (e.g., Fig.~\ref{fig:planktonsim}c), and use GP to map the \emph{log-measurements}. As a result,
the entropy criterion (\ref{eq:1a}) has to be optimized in the transformed log-scale.
%If we want to predict the actual measurement at an unobserved location, say $u$, using the posterior data $d_n$, it may seem appropriate to use the best unbiased predictor of the log-measurement $z_u$, $\mu_{z_u \mid d_{n}}$ (\ref{eq:6}), to compose the predictor $\exp\{\mu_{z_u \mid d_{n}}\}$\footnote{If no posterior data $d_n$ is available, the predictor $\exp\{\mu_{z_u \mid d_{n}}\}$ reduces to $\exp\{\mu_{z_u}\}$.} for predicting the actual measurement.
%This predictor is, however, biased as it underestimates the expected value of the actual measurement.
%We will defer the description of an unbiased predictor to the next section (specifically, equation (\ref{eq:10})).

We will apply $i$MASP($1$) to sampling GP and determine if $\pi^{1}$ exhibits adaptive and hotspot sampling properties.
Let $\{Z_x\}_{x \in \set{X}}$ denote a GP, i.e., the joint distribution over any finite subset of $\{Z_x\}_{x \in \set{X}}$ is Gaussian \cite{Rasmussen06}.
The GP can be completely specified by its mean
$\mu_{Z_x} \defeq \mathbb{E}[Z_x]$ and covariance
$\sigma_{Z_x Z_{u}} \defeq \mbox{cov}[Z_x, Z_{u}]$
for $x, u \in \set{X}$.
We adopt a common assumption that the GP is second-order stationary, i.e., it has a constant mean and a stationary covariance structure
(i.e., $\sigma_{Z_x Z_{u}}$ is a function of $x - u$ for all $x, u \in \set{X}$).
In this paper, we assume that the mean and covariance structure of $Z_x$ are known. 
Given $d_{n}$, the distribution of $Z_x$ is Gaussian with posterior mean and covariance
\vspace{-2mm}
\begin{equation}
\displaystyle 
\mu_{Z_x \mid d_{n}}
= \mu_{Z_x} + \Sigma_{x \ve{x}_{0:n}} \Sigma^{-1}_{\ve{x}_{0:n} \ve{x}_{0:n}} \{ \ve{z}_{\ve{x}_{0:n}} - \bm{\mu}_{\ve{Z}_{\ve{x}_{0:n}}} \}^{\top}\vspace{-1mm}
\label{eq:6}
\end{equation}
\begin{equation}
\displaystyle \sigma_{Z_x Z_u \mid d_{n}} = \sigma_{Z_x Z_u} - \Sigma_{x \ve{x}_{0:n}} \Sigma^{-1}_{\ve{x}_{0:n} \ve{x}_{0:n}} \Sigma_{\ve{x}_{0:n} u}%\vspace{-0.5mm}
\label{eq:7}
\end{equation}
where, for every pair of locations $v, w$ of $\ve{x}_{0:n}$,
$\bm{\mu}_{\ve{Z}_{\ve{x}_{0:n}}}$
is a row vector with mean components $\mu_{Z_{v}}$,
$\Sigma_{x \ve{x}_{0:n}}$ is a row vector with covariance components $\sigma_{Z_{x}Z_{v}}$,
$\Sigma_{\ve{x}_{0:n} u}$ is a column vector with covariance components $\sigma_{Z_{v}Z_{u}}$,
and $\Sigma_{\ve{x}_{0:n} \ve{x}_{0:n}}$ is a covariance matrix with components $\sigma_{Z_{v}Z_{w}}$.
%Note that the posterior mean $\mu_{z_{x}|d_{n}}$ (\ref{eq:6}) corresponds to the optimal predictor $\hat{z}_x(d_{n})$ for predicting the measurement $z_{x}$ at the unobserved location $x$. 
An important property of $\sigma_{Z_x Z_u \mid d_{n}}$ is its independence of $\ve{z}_{\ve{x}_{1:n}}$.

Policy $\pi^{1}$ can be reduced to be \emph{non-adaptive}: observe that each stagewise reward is independent of the measurements\vspace{-1mm}
\begin{equation}
\mathbb{H}[\ve{Z}_{\tau(\ve{x}_{i}, \ve{a}_{i})}| d_{i}] = \log\sqrt{(2\pi e)^k \ |\Sigma_{\ve{Z}_{\tau(\ve{x}_{i}, \ve{a}_{i})}\mid d_i}|}\vspace{-1mm}
\label{eq:7a}
\end{equation}
where $\Sigma_{\ve{Z}_{\tau(\ve{x}_{i}, \ve{a}_{i})}\mid d_i}$ is a covariance matrix with components $\sigma_{Z_{x}Z_{u}\mid d_i}$, $x, u$ of $\tau(\ve{x}_{i}, \ve{a}_{i})$, that are independent of
$\ve{z}_{\ve{x}_{1:n}}$.
As a result, it follows from (\ref{eq:4r}) that
$U^{\pi^1}_{i}(d_{i})$ and $\pi^{1}_{i}(d_{i})$ are independent of $\ve{z}_{\ve{x}_{1:n}}$ for $i = 0, \ldots, n-1$.
%Hence, $\pi^{1}$ is non-adaptive.
The expectations in $i$MASP($1$) (\ref{eq:4r}) can then be integrated out.
% to reduce $U^{\pi^1}_0(d_0)$ to $U^{\pi^n}_{0}(d_{0})$. So, $\pi^{1}$ offers no performance advantage over $\pi^{n}$. 
%Since $U^{\pi^1}_{i}(d_{i})$ is independent of $\ve{z}_{\ve{x}_{1:n}}$, the expectations in $i$MASP($1$) (\ref{eq:4r}) can be integrated out. 
As a result, $i$MASP($1$) for sampling GP can be reduced to a 1-stage deterministic problem 
$U^{\pi^1}_0(d_0) =$
$\sum^{n-1}_{i=0} \displaystyle\max_{\ve{a}_i} \mathbb{H}[\ve{Z}_{\tau(\ve{x}_{i}, \ve{a}_{i})}| d_{i}]
=$ $\displaystyle\max_{\ve{a}_0, \ldots, \ve{a}_{n-1}}$ $\sum^{n-1}_{i=0}\mathbb{H}[\ve{Z}_{\tau(\ve{x}_{i}, \ve{a}_{i})}| d_{i}]
= \displaystyle\max_{\ve{a}_{0:n-1}}$ $\mathbb{H}[\ve{Z}_{\tau(\ve{x}_{0:n-1}, \ve{a}_{0:n-1})}| d_{0}]
= U^{\pi^n}_{0}(d_{0})$.
This indicates the induced optimal values from solving $i$MASP($1$) and $i$MASP($n$) are equal. 
So, $\pi^{1}$ offers no performance advantage over $\pi^{n}$. 

Based on the above analysis, the following sufficient conditions, when met, guarantee that adaptivity has no benefit under an assumed environmental model:\vspace{-0.5mm}
\begin{theorem}
If $\mathbb{H}[\ve{Z}_{\tau(\ve{x}_{i}, \ve{a}_{i})}|d_{i}]$ is independent of
$\ve{z}_{\ve{x}_{1:n}}$ for stage $i = 0, \ldots, n-1$, $i$MASP($1$) and $\pi^{1}$ can be reduced to be single-staged and non-adaptive, respectively. \vspace{-0.5mm}
\label{thm:2}
\end{theorem}
For example, Theorem~\ref{thm:2} also holds for the simple case of an \emph{occupancy grid map} modeling an obstacle-ridden environment, which typically assumes $z_x$ for $x \in \set{X}$ to be independent. As a result, 
$\mathbb{H}[\ve{Z}_{\tau(\ve{x}_{i}, \ve{a}_{i})}|d_{i}]$ can be reduced to a sum of prior entropies over the unobserved locations $\tau(\ve{x}_{i}, \ve{a}_{i})$, which are independent of $\ve{z}_{\ve{x}_{1:n}}$.

Policy $\pi^{1}$ performs \emph{wide-area coverage} only: to maximize stagewise rewards (\ref{eq:7a}), 
$\pi^{1}$ selects new locations with large posterior variance for observation.
%/observes new locations with large posterior variance, which can be found in sparsely sampled areas. 
%However, it does not account for maximization of hotspot sampling.\vspace{2mm}\\
If we assume isotropic covariance structure (i.e., the covariance $\sigma_{Z_x Z_{u}}$ decreases monotonically with $||x-u||$) \cite{Rasmussen06},
the posterior data $d_i$ provide the least amount of information on unobserved locations that are far away from all observed locations.
As a result, the posterior variance of unobserved locations in sparsely sampled regions are still largely unreduced by the posterior data $d_i$ from the observed locations.
Hence, by exploring the sparsely sampled areas, a large expected entropy can be absorbed into the selected observation paths. 
%Realize that it does not account for maximization of hotspot sampling.
%{\bf Log-Gaussian Process}. For sampling GP, the induced exploration policy $\pi^1$ from solving MASP($1$) is optimal 
%in the sense that it maximizes the entropy of the observation paths 
%with respect to the log-measurements of the hotspot field. As shown previously, $\pi^1$ is non-adaptive and performs only wide-area coverage for GP. 
Using the observations selected from wide-area coverage, the field of \emph{original} measurements may not be mapped well because the under-sampled hotspots with extreme, highly-varying measurements contribute considerably to map entropy in the original scale, as discussed below.\vspace{1mm}

\noindent
{\bf Log-Gaussian Process}.
To map the original, rather than the log-, measurements directly, it is a conventional practice in geostatistics to use the $\ell$GP. Consequently, the entropy criterion (\ref{eq:1a}) is optimized in the original scale. To do this, let $\{Y_x\}_{x \in \set{X}}$ denote a $\ell$GP: if $Z_x = \log Y_x$, $\{Z_x\}_{x \in \set{X}}$ is a GP. 
So, the positive-valued $Y_x=\exp\{ Z_x\}$ denotes the original random measurement at location $x$.
It is straightforward to derive the predictive properties of $\ell$GP from that of GP as shown in \cite{LowAAMAS08}.

A $\ell$GP can model a field with hotspots that exhibit much higher spatial variability
than the rest of the field: 
%We will also show that for sampling $\ell$GP, $\pi^1$ is adaptive, and performs both wide-area coverage and hotspot sampling.
\begin{figure}
\begin{tabular}{ccc}
\multirow{3}{*}[1.35cm]{\hspace{-3mm}\epsfig{figure=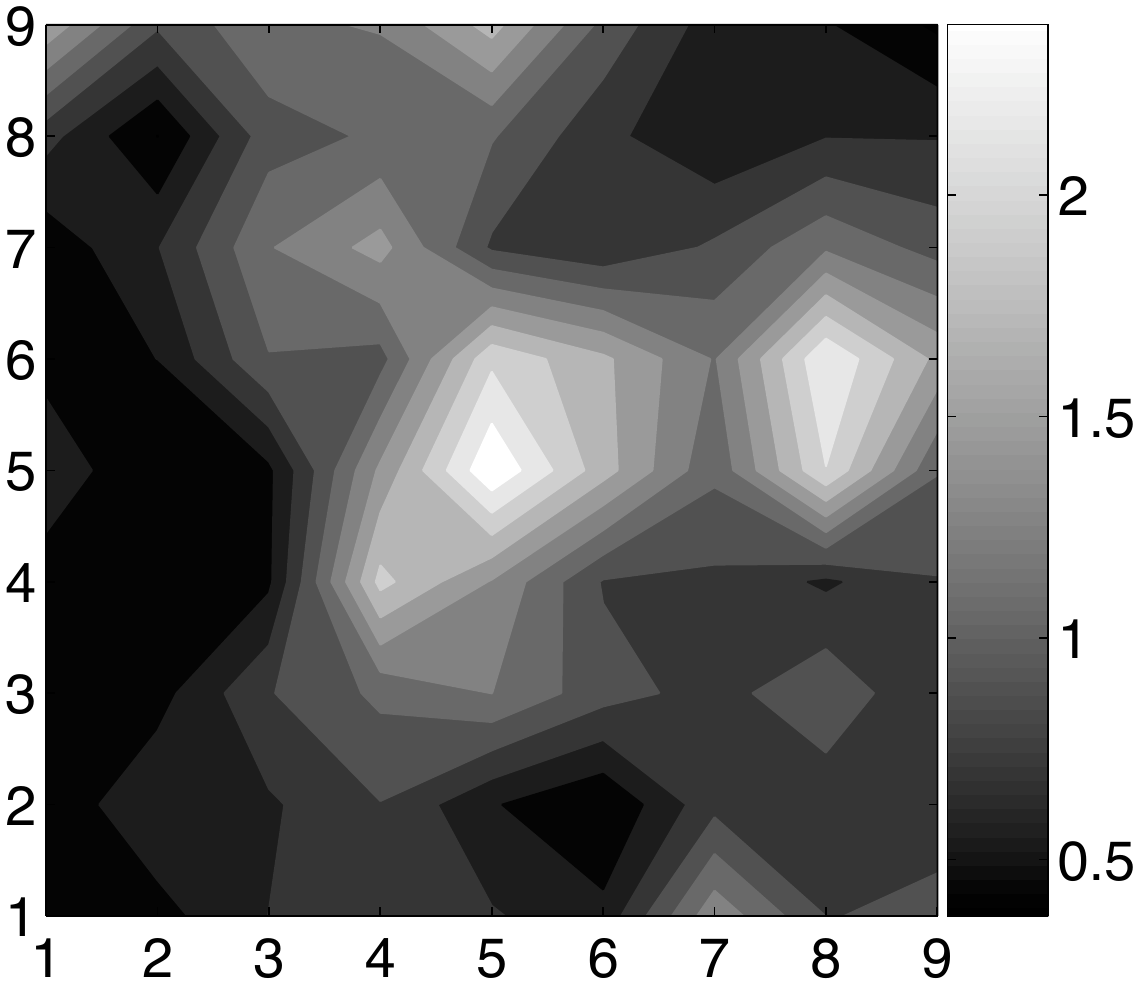,width=3.5cm}} & \hspace{-4mm}\epsfig{figure=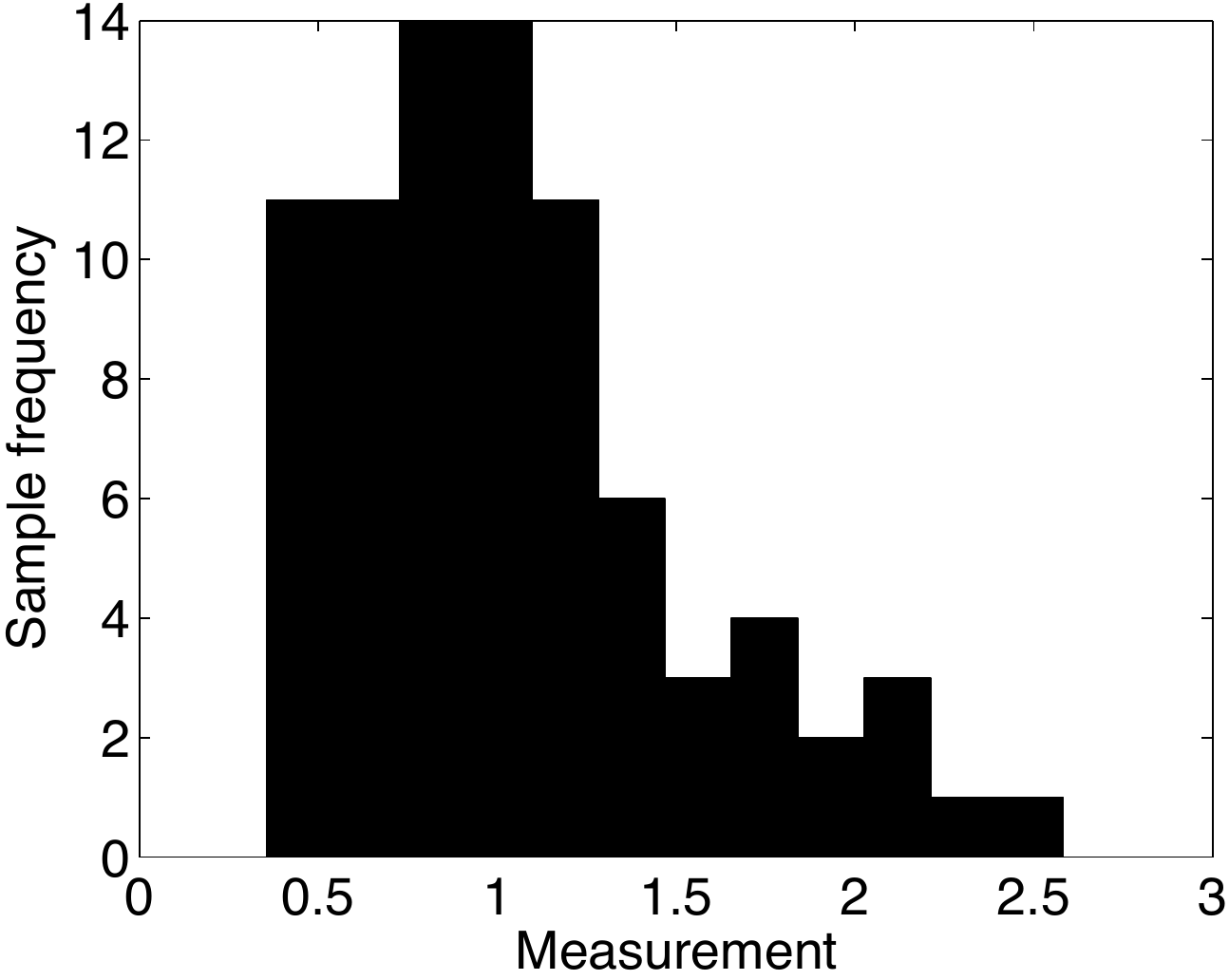,width=2cm} &
\multirow{3}{*}[1.35cm]{\hspace{-3mm}\epsfig{figure=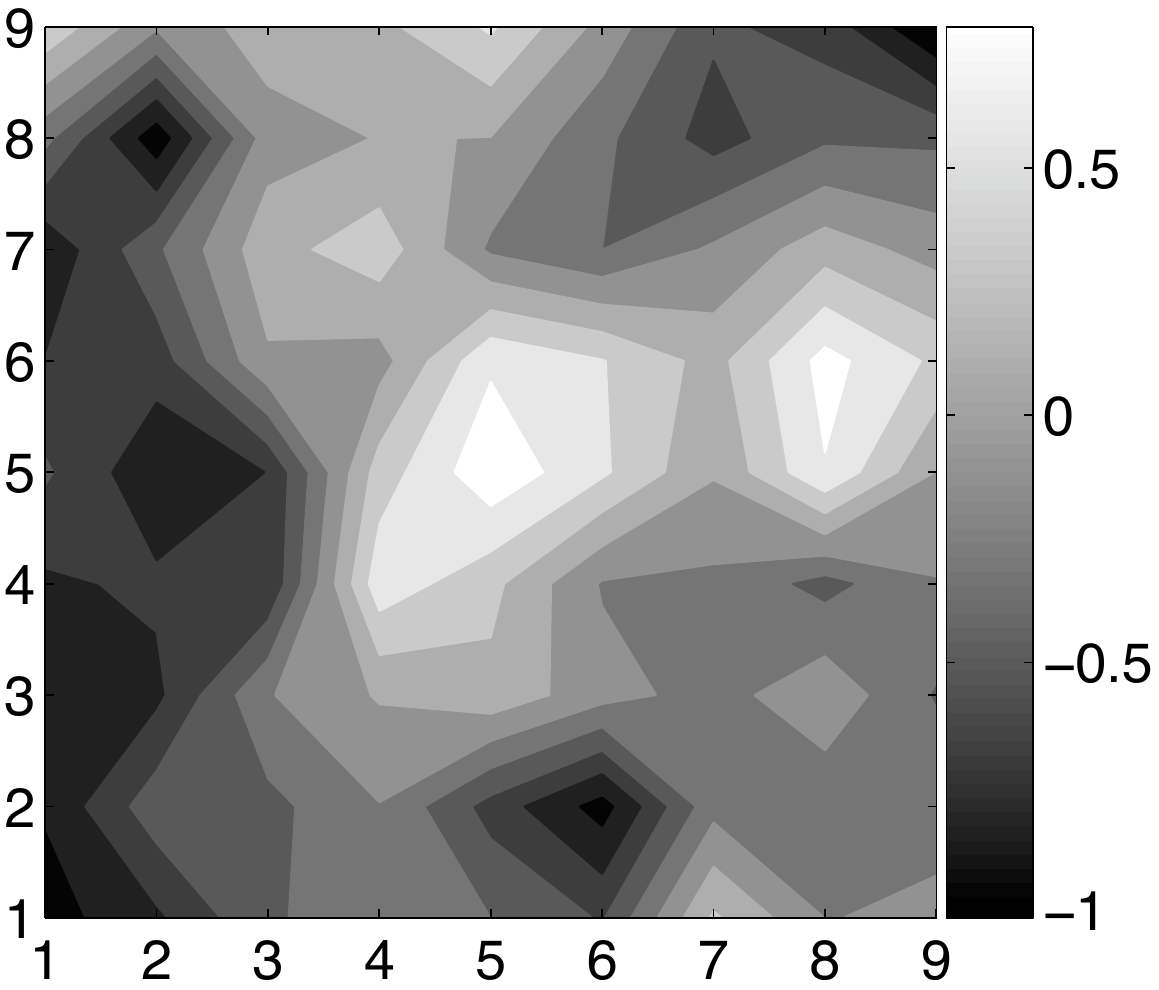,width=3.5cm}}\\
& \multirow{1}{*}[1.5cm]{\hspace{8mm}\footnotesize(b)} &\vspace{-4mm}\\
& \hspace{-4mm}\epsfig{figure=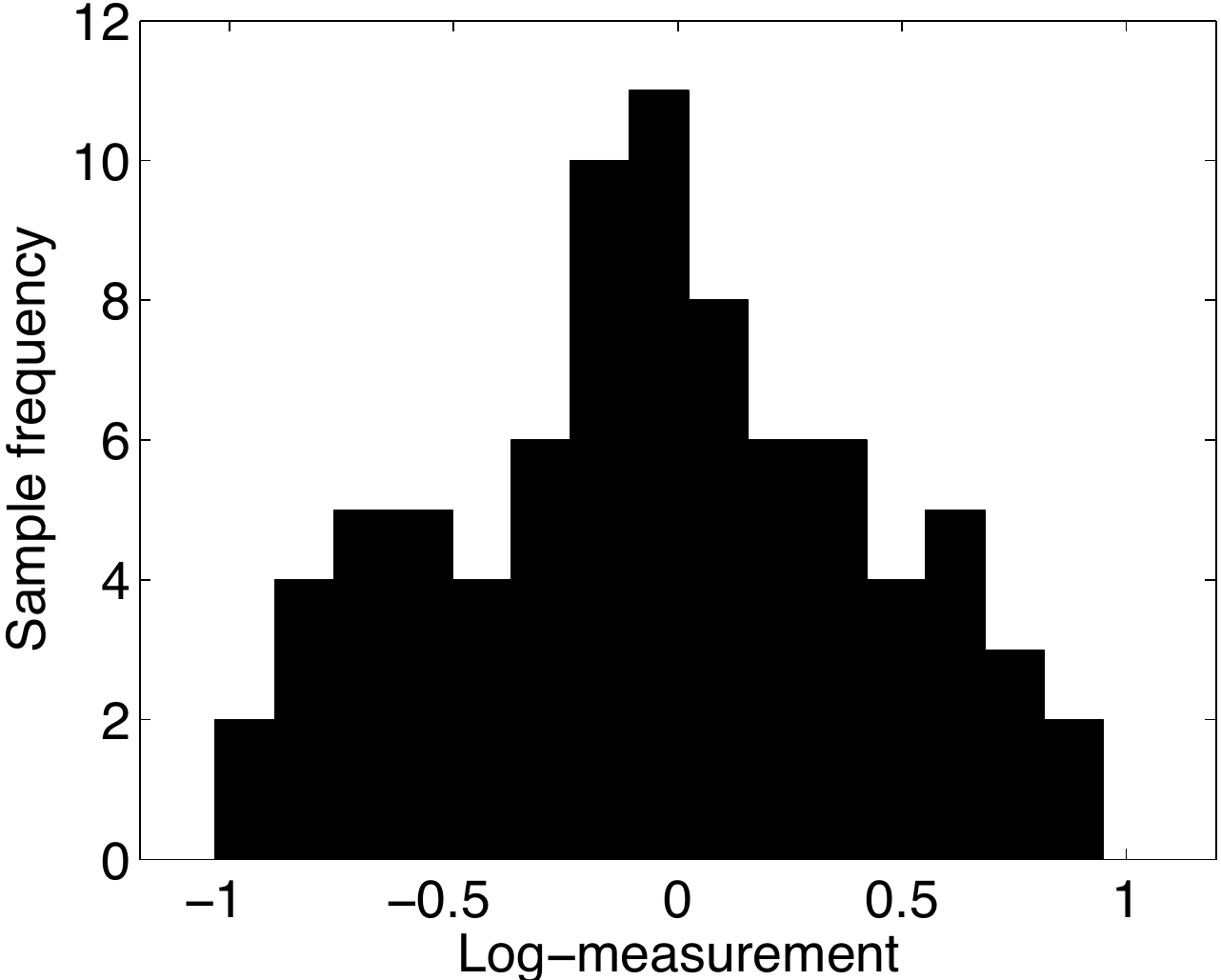,width=2cm} & \vspace{-5mm}\\
\hspace{-4mm}{\footnotesize(a)} & \multirow{1}{*}[1cm]{\hspace{8mm}(c)} & \hspace{-5mm}{\footnotesize(d)}\\
\end{tabular}\vspace{-4mm}
\caption{Hotspot field simulation: (a) $\ell$GP and (d) GP realizations with their 1D sample frequency distributions shown, respectively, in (b) and (c).}\vspace{-4mm}
\label{fig:planktonsim}
\end{figure}
%\begin{figure}
%\begin{tabular}{cccc}
%\multicolumn{2}{c}{\hspace{-1mm}\epsfig{figure=lgpsim,width=4cm}} & \multicolumn{2}{c}{\hspace{-1mm}\epsfig{figure=gpsim,width=4cm}}\vspace{-1mm}\\
%\hspace{-2mm}
%\epsfig{figure=lgphistsim,width=3cm} & \multirow{1}{*}[1.2cm]{\hspace{-5mm}(a)} & \hspace{-1mm}\epsfig{figure=gphistsim,width=3cm} & \multirow{1}{*}[1.2cm]{\hspace{-4mm}(b)} \vspace{-4mm}\\
%\end{tabular}
%\caption{Hotspot field simulation via the (a) $\ell$GP, and (b) GP.}\vspace{-4mm}
%\label{fig:planktonsim}
%\end{figure}
%So, the positive-valued $y_x = \exp\{z_x\}$ denotes the actual measurement at location $x$. Properties of the prior and posterior means and covariances for $\ell$GP can be found in \cite{LowAAMAS08}. 
Figs.~\ref{fig:planktonsim}a and~\ref{fig:planktonsim}d compare the realizations of $\ell$GP and GP; the GP realization results from taking the log of the $\ell$GP measurements. This does not just dampen the extreme measurements, but also dampens and amplifies the difference between extreme and small measurements respectively, thus removing the positive skew (compare Figs.~\ref{fig:planktonsim}b and~\ref{fig:planktonsim}c).
Compared to the GP realization, the $\ell$GP one thus exhibits higher spatial variability within hotspots but lower variability in the rest of the field. This intuitively explains why wide-area coverage suffices for GP but hotspot sampling is further needed for $\ell$GP.

% $\ell$GP has the mean $\mu_{y_x}\stackrel{\rm def}{=} \mathbb{E}[y_x] = \exp\{ \mu_{z_x} + \sigma_{z_x z_x}/2 \}$ and covariance
%$\sigma_{y_x y_{u}}\stackrel{\rm def}{=} \mbox{cov}[y_x, y_{u}] = \mu_{y_x}\mu_{y_{u}}(\exp\{ \sigma_{z_x z_{u}} \} - 1)$ 
%for $x, u \in \set{X}$.
%We know that the distribution of $z_x$ given the posterior data $d_{n}$ is Gaussian. 
%Since the transformation from $\ve{z}_{\ve{x}_{0:n}}$ to $\ve{y}_{\ve{x}_{0:n}}$ is invertible, 
%the distribution of $y_x$ given the posterior data $d_{n}$ is log-Gaussian with the posterior mean
%$
%\displaystyle \mu_{y_x \mid d_{n}} = \exp\{\mu_{z_x \mid d_{n}} + \sigma^2_{z_x \mid d_{n}}/2 \}
%$  and variance
%$
%\displaystyle \sigma^2_{y_x \mid d_{n}} = \mu^2_{y_x \mid d_{n}} (\exp\{\sigma^2_{z_x \mid d_{n}}\} - 1)
%$ where $\mu_{z_x|d_{n}}$ and $\sigma^2_{z_x|d_{n}}$ are the Gaussian posterior mean (\ref{eq:6}) and variance (\ref{eq:7}) respectively.

Policy $\pi^{1}$ is \emph{adaptive}: observe that each stagewise reward depends on the previously sampled data $d_i$:\vspace{-2mm}
\begin{equation}
\mathbb{H}[\ve{Y}_{\tau(\ve{x}_{i}, \ve{a}_{i})}| d_{i}] = \log\sqrt{(2\pi e)^k |\Sigma_{\ve{Z}_{\tau(\ve{x}_{i}, \ve{a}_{i})}\mid d_i}|} + \bm{\mu}_{\ve{Z}_{\tau(\ve{x}_{i}, \ve{a}_{i})}|d_i} \ve{1}^{\top} \vspace{-2mm}
\label{eq:11a}
\end{equation}
where $\bm{\mu}_{\ve{Z}_{\tau(\ve{x}_{i}, \ve{a}_{i})}|d_i}$ is a mean vector with components $\mu_{Z_x|d_i}$ for $x$ of $\tau(\ve{x}_{i}, \ve{a}_{i})$. Since $\mu_{Z_x|d_i}$ depends on $d_i$ by (\ref{eq:6}), $\mathbb{H}[\ve{Y}_{\tau(\ve{x}_{i}, \ve{a}_{i})}| d_{i}]$ depends on $d_i$. Consequently, it follows from (\ref{eq:4r}) that
$U^{\pi^1}_{i}(d_{i})$ and $\pi^{1}_{i}(d_{i})$ depend on $d_i$ for $i = 0, \ldots, n-1$. Hence, $\pi^{1}$ is \emph{adaptive}. 

Policy $\pi^{1}$ performs both \emph{hotspot sampling} and \emph{wide-area coverage}:
to maximize stagewise rewards (\ref{eq:11a}), $\pi^{1}$ selects new locations with large Gaussian posterior variance and mean for observation.
So, it directs exploration towards sparsely sampled areas and hotspots.\vspace{-2mm}
\section{Value-Function Approximations}
\label{sect:vfa}
%\vspace{-1mm}
\noindent
{\bf Strictly Adaptive Exploration}.
With a team of $k>1$ robots, $\pi^{1}$ collects $k>1$ observations per stage, thus becoming \emph{partially adaptive}. We will now derive the optimal \emph{strictly adaptive} policy (in particular, for sampling $\ell$GP),
which, among policies of all adaptivity, selects paths with the largest expected entropy.
By Def.~\ref{def:2}, a strictly adaptive policy has to be structured to collect only $1$ observation per stage. 
%To achieve this, we can restrict, in each stage, only one robot in the team to observe a new location while the other robots stay put. 
To achieve strict adaptivity, $i$MASP($1$) (\ref{eq:4r}) can be revised as follows:
(a) The space $\set{A}(\ve{x}_i)$ of simultaneous joint actions is reduced to a constrained set $\set{A}'(\ve{x}_i)$ of joint actions that allows one robot to move to observe a new location and the other robots stay put.
This tradeoff for strict adaptivity allows $\set{A}'(\ve{x}_i)$ to grow linearly, rather than exponentially, with the number of robots;
(b) We constrain each robot to explore a path of at most $n$ new adjacent locations; this can be viewed as an energy consumption constraint on each robot. The horizon then spans $k\times n$, rather than $n$, stages, which reflects the additional time of exploration incurred by strict adaptivity;
(c) If $\ve{a}_{i}\in \set{A}'(\ve{x}_i)$, the assignment $\ve{x}_{i+1} \leftarrow \tau(\ve{x}_{i}, \ve{a}_{i})$ moves one chosen robot to a new location $x_{i+1}$ while the other unselected robots stay put at their current locations. 
Then, only one component of $\ve{x}_{i}$ is changed to $x_{i+1}$ to form $\ve{x}_{i+1}$; the other components of $\ve{x}_{i+1}$ are unchanged from $\ve{x}_{i}$.
Hence, there is only one unobserved component $Y_{x_{i+1}}$ in $\ve{Y}_{\ve{x}_{i+1}}$; the other components of $\ve{Y}_{\ve{x}_{i+1}}$ are already observed in the previous stages and can be found in $d_i$.
As a result, the probability distribution of $\ve{Y}_{\ve{x}_{i+1}}$ can be simplified to a univariate $Y_{x_{i+1}}$.

These revisions of $i$MASP($1$) yield the strictly adaptive exploration problem called $i$MASP($\frac{1}{k}$):
%which resolves to the following $kn$-stage dynamic programming equations:
\vspace{-2mm}
\begin{equation}
\hspace{-2mm}
\begin{array}{rl}
\displaystyle U_{i}(d_{i}) =& \hspace{-2mm}\displaystyle\max_{\ve{a}_{i} \in \set{A}'(\ve{x}_{i})}
\mathbb{H}[Y_{x_{i+1}}|d_i] + \displaystyle\int f(y_{x_{i+1}}|d_i) \ U_{i+1}(d_{i+1}) \ \mbox{d}y_{x_{i+1}}\\
=& \hspace{-2mm}\displaystyle\max_{\ve{a}_{i} \in \set{A}'(\ve{x}_i)} \mathbb{H}[Y_{x_{i+1}}|d_i] + \mathbb{E}[ U_{i+1}(d_i, x_{i+1}, Y_{x_{i+1}})|d_i ]\\
\displaystyle U_{t}(d_t) =& \hspace{-2mm}
\displaystyle\max_{\ve{a}_{t} \in \set{A}'(\ve{x}_t)}
\mathbb{H}[Y_{x_{t+1}}| d_t]\vspace{-4mm}
\end{array}
\label{eq:4r4}
\end{equation}
for stage $i = 0,\ldots,t-1$ where $t=kn-1$. Without ambiguity, we omit the superscript $\pi^{\frac{1}{k}}$ (i.e., the optimal strictly adaptive policy) from the optimal value functions above. 
%The optimal strictly adaptive policy $\pi^{\frac{1}{k}} = \langle\pi^{\frac{1}{k}}_0(d_0),\ldots, \pi^{\frac{1}{k}}_{t}(d_{t})\rangle$ is produced by solving $i$MASP($\frac{1}{k}$).

Since $Y_{x_{i+1}}$ is continuous, it entails infinite state transitions.
So, $\mathbb{E}[ U_{i+1}(d_i, x_{i+1}, Y_{x_{i+1}})| d_i ]$ has to be evaluated in closed form for $i$MASP($\frac{1}{k}$) to be solved exactly. This can be performed for $t = 1$. When $t > 1$,
the expectation of the optimal value function results in an integral that is too complex to be evaluated.
%we are not aware of any computationally feasible methods to solve $i$MASP($\frac{1}{k}$) with  exactly.
Hence, we will resort to approximating $i$MASP($\frac{1}{k}$) as described below.
For ease of exposition, we will revert to using $Z_{x_{i+1}} = \log Y_{x_{i+1}}$ for $\ell$GP from now on.\vspace{1mm}\\
{\bf Approximately Optimal Exploration}. 
To approximate $i$MASP($\frac{1}{k}$), we will first approximate the expectation in (\ref{eq:4r4}) from below and above using the ${\nu}$-fold generalized Jensen and Edmundson-Madansky (EM) bounds respectively \cite{Huang77}.
To do this, we need the following convexity result for $i$MASP($\frac{1}{k}$) (\ref{eq:4r4}):
\begin{lemma}
$U_i(d_{i})$ is convex in $\ve{z}_{\ve{x}_{0:i}}$ for $i = 0, \ldots, t$.\vspace{0mm}
\label{lem:3}
\end{lemma}
%To do this, we claim that $U_{i+1}(d_i, x_{i+1}, z_{x_{i+1}})$ is convex in $z_{x_{i+1}}$. 
Let the support of $Z_{x_{i+1}}$ given $d_i$ be $\set{Z}^{\nu}_{x_{i+1}}$ that is partitioned into $\nu$ disjoint intervals $\set{Z}^{[j]}_{x_{i+1}}=[\overline{z}^{[j-1]}_{x_{i+1}},\overline{z}^{[j]}_{x_{i+1}}]$ for $j = 1,\ldots, \nu$.
Then, 
%the expectation can be approximated from below:
\vspace{-2mm}
\begin{equation}
\hspace{-2mm}
\begin{array}{rl}
\displaystyle\sum^{\nu}_{j=1} \underline{p}^{[j]}_{x_{i+1}} U_{i+1}(d_i, x_{i+1}, \underline{z}^{[j]}_{x_{i+1}}) &\hspace{-2mm}\leq \mathbb{E}[U_{i+1}(d_i, x_{i+1}, Z_{x_{i+1}})| d_i]\vspace{-3mm}\\
&\hspace{-2mm}\leq \displaystyle\sum^{\nu}_{j=0} \overline{p}^{[j]}_{x_{i+1}} U_{i+1}(d_i, x_{i+1}, \overline{z}^{[j]}_{x_{i+1}})\vspace{-3mm}
\end{array}
\label{eq:5v}\vspace{-2mm}
\end{equation}
where $\underline{p}^{[j]}_{x_{i+1}}\defeq \displaystyle\int_{\set{Z}^{[j]}_{x_{i+1}}}f(z_{x_{i+1}}| d_i) \ \mbox{d}z_{x_{i+1}}$ and\vspace{-1mm}\\
$\underline{z}^{[j]}_{x_{i+1}} \hspace{-1mm}\defeq\hspace{-1mm}\displaystyle \frac{1}{\underline{p}^{[j]}_{x_{i+1}}}\hspace{-1.2mm}\int_{\set{Z}^{[j]}_{x_{i+1}}}\hspace{-3mm} z_{x_{i+1}} f(z_{x_{i+1}}|d_i) \ \mbox{d}z_{x_{i+1}}\vspace{-1mm}$ for $j=1,\ldots,\nu$,\vspace{1mm}\\
$\overline{p}^{[j]}_{x_{i+1}} \defeq \displaystyle \underline{p}^{[j]}_{x_{i+1}} \frac{\underline{z}^{[j]}_{x_{i+1}} - \overline{z}^{[j-1]}_{x_{i+1}}}{\overline{z}^{[j]}_{x_{i+1}} - \overline{z}^{[j-1]}_{x_{i+1}}} + \underline{p}^{[j+1]}_{x_{i+1}}\frac{\overline{z}^{[j+1]}_{x_{i+1}}-\underline{z}^{[j+1]}_{x_{i+1}}}{\overline{z}^{[j+1]}_{x_{i+1}}-\overline{z}^{[j]}_{x_{i+1}}}$ for $j=0,\ldots,\nu$, and
$\underline{p}^{[0]}_{x_{i+1}} \hspace{-2mm}:= \underline{p}^{[\nu+1]}_{x_{i+1}}\hspace{-1mm} :=\underline{z}^{[0]}_{x_{i+1}} \hspace{-2mm}:= \underline{z}^{[\nu+1]}_{x_{i+1}}\hspace{-1mm} := \overline{z}^{[-1]}_{x_{i+1}} \hspace{-1mm}:=0$.
By increasing $\nu$ to refine the partition, the bounds can be improved.	
%The generalized Jensen bounds (\ref{eq:5v}) can be viewed as approximating the continuous state variable $Z_{x_{i+1}}$ using a discrete one with a distribution at points $\underline{z}^{[j]}_{x_{i+1}}$ of probability $\underline{p}^{[j]}_{x_{i+1}} > 0$ for $j = 1, \ldots, \nu$ where $\sum^{\nu}_{j=1} \underline{p}^{[j]}_{x_{i+1}} = 1$. This is similarly true for the EM bound.

The upper approximate problem $\overline{i}$MASP($\frac{1}{k}$) can be constructed from $i$MASP($\frac{1}{k}$) (\ref{eq:4r4}) by replacing the expectation with the upper EM bound (\ref{eq:5v})
to yield the optimal value functions $\overline{U}^{\nu}_{i}(d_i)$ for $i=0,\ldots,t$.
%:\vspace{-2mm}
Similarly, the lower approximate problem $\underline{i}$MASP($\frac{1}{k}$) can be constructed from $i$MASP($\frac{1}{k}$) (\ref{eq:4r4}) by replacing the expectation with the lower Jensen bound (\ref{eq:5v}) to yield the optimal value functions $\underline{U}^{\nu}_{i}(d_i)$ for $i=0,\ldots,t$ and optimal policy $\underline{\pi}^{\frac{1}{k}}$.

The next result uses the induced optimal values from solving the lower and upper approximate problems to monotonically bound the maximum expected entropy achieved by the optimal strictly adaptive policy $\pi^{\frac{1}{k}}$:\vspace{-0mm}
%To show this, we make use of a stronger convexity result for the optimal value functions of $i$MASP($\frac{1}{k}$):
%\begin{lemma}
%$U_i(d_{i})$ is convex in $\ve{z}_{\ve{x}_{0:i}}$ for $i = 0, \ldots, t$.
%\label{lem:3}
%\end{lemma}
\begin{theorem}
If $\set{Z}^{\nu+1}_{x_{i+1}}$ is obtained by splitting one of the intervals in $\set{Z}^{\nu}_{x_{i+1}}$,
$\underline{U}^{\nu}_{i}(d_i) \leq \underline{U}^{\nu + 1}_{i}(d_i) \leq U_{i}(d_i) \leq \overline{U}^{\nu + 1}_{i}(d_i) \leq \overline{U}^{\nu}_{i}(d_i)$ for $i = 0, \ldots, t$.\vspace{-0.5mm}
\label{thm:9}
\end{theorem}
A previous result of \citeauthor{LowAAMAS08} \shortcite{LowAAMAS08} has guaranteed that $\underline{\pi}^{\frac{1}{k}}$ can achieve an expected entropy not worse than $\underline{U}^{\nu}_0(d_0)$.
But, that result does not account for how much it differs from
the maximum expected entropy achieved by $\pi^{\frac{1}{k}}$.
With the upper bound of Theorem~\ref{thm:9}, this error difference can be bounded:
\begin{corollary}
$\underline{\pi}^{\frac{1}{k}}$ is guaranteed to achieve an expected entropy that is not more than $\overline{U}^{\nu}_{0}(d_0) - \underline{U}^{\nu}_{0}(d_0)$ from the maximum expected entropy $U_{0}(d_0)$ achieved by $\pi^{\frac{1}{k}}$.%\vspace{1mm}
\label{col:2}
\end{corollary}
{\bf Bounds on Performance Advantage of Adaptive Exploration}.
A previous result of \citeauthor{LowAAMAS08} \shortcite{LowAAMAS08} has established the performance advantage of optimal adaptive over non-adaptive policies.
%In terms of minimizing an optimizing criterion such as the expected posterior map error \cite{LowAAMAS08} or entropy, the optimal adaptive policy $\pi^1$ performs better or, if not, at least as well as the optimal non-adaptive policy $\pi^n$ as shown in \cite{LowAAMAS08}. This result can be generalized to cover the entire adaptivity spectrum: by increasing the adaptivity of the exploration problem, the performance of the induced optimal policy improves monotonically. In a later section, we will show how this performance advantage can be quantified, which is not described in \cite{LowAAMAS08}.
Realizing the extent of such an advantage is important if adaptivity incurs a cost.
In particular, we are interested in quantifying the performance difference between the strictly adaptive $\pi^{\frac{1}{k}}$ and the non-adaptive $\pi^n$.
This performance advantage of $\pi^{\frac{1}{k}}$ over $\pi^n$ is defined as the difference of their achieved maximum expected entropies $U_{0}(d_{0}) - U^{\pi^n}_{0}(d_{0})$.
Using the induced optimal values from solving the approximate problems (Theorem~\ref{thm:9}), the advantage $U_{0}(d_{0}) - U^{\pi^n}_{0}(d_{0})$ can be bounded between $\underline{U}^{\nu}_{0}(d_0) - U^{\pi^n}_{0}(d_{0})$ and $\overline{U}^{\nu}_{0}(d_0) - U^{\pi^n}_{0}(d_{0})$.
A large lower bound $\underline{U}^{\nu}_{0}(d_0) - U^{\pi^n}_{0}(d_{0})$ implies $\pi^{\frac{1}{k}}$ is to be preferred. A small upper bound $\overline{U}^{\nu}_{0}(d_0) - U^{\pi^n}_{0}(d_{0})$ implies $\pi^{n}$ performs close to that of $\pi^{\frac{1}{k}}$ and should be preferred if it is more costly to deploy $\pi^{\frac{1}{k}}$.
For GP, this advantage is zero as $\pi^{\frac{1}{k}}$ can be reduced to be non-adaptive as shown previously.\vspace{1mm}\\ 
{\bf Real-Time Dynamic Programming}.
For our bounding approximation scheme, the state size grows exponentially with the number of stages. 
This is due to the nature of dynamic programming problems (e.g., $\underline{i}$MASP($\frac{1}{k}$)), which takes into account all possible states. 
To alleviate this computational difficulty, we modify the anytime algorithm URTDP of \citeauthor{LowAAMAS08} \shortcite{LowAAMAS08} based on $\underline{i}$MASP($\frac{1}{k}$), which can guarantee its policy performance in real time.
It simulates greedy exploration paths through a large state space, resulting in desirable properties of focused search and good anytime behavior. The greedy exploration is guided by computationally efficient, informed initial heuristic bounds independent of state size.%\vspace{-3mm}

In URTDP (Algorithm~\ref{alg2}), each simulated path involves an alternating selection of actions and their corresponding outcomes till the last stage. Each action is selected based on the upper bound (line~\ref{line:urtdp0}). 
For each encountered state, the algorithm maintains both lower and upper bounds, which are used to derive the uncertainty of its corresponding optimal value function. It exploits them to guide future searches in an informed manner; it explores the next state/outcome with the greatest amount of uncertainty (lines~\ref{line:urtdp1}-\ref{line:urtdp4}). 
Then, the algorithm backtracks up the path to update the upper heuristic bounds using
$\max_{\ve{a}_i} \overline{Q}_i(\ve{a}_i, d_i)$ (line~\ref{line:urtdp2}) where\vspace{-1.9mm}
$$\overline{Q}_i(\ve{a}_i, d_i) \defeq \mathbb{H}[Y_{x_{i+1}}|d_i] +
\sum^{\nu}_{j=1} \underline{p}^{[ j]}_{x_{i+1}} 
\overline{U}_{i+1}(d_i,  x_{i+1}, \underline{z}^{[ j]}_{x_{i+1}})\vspace{-2mm}$$
and the lower bounds via 
$\max_{\ve{a}_i} \underline{Q}_i(\ve{a}_i, d_i)$ (line~\ref{line:urtdp3}) where\vspace{-2.3mm}
$$\underline{Q}_i(\ve{a}_i, d_i) \defeq \mathbb{H}[Y_{x_{i+1}}|d_i] +
\sum^{\nu}_{j=1} \underline{p}^{[ j]}_{x_{i+1}}
\underline{U}_{i+1}(d_i, x_{i+1}, \underline{z}^{[ j]}_{x_{i+1}}) \ .\vspace{-2mm}$$ 
When an exploration policy is requested, we provide the greedy policy induced by the lower bound. The policy performance has a similar guarantee to Corollary~\ref{col:2}.
%that of $\underline{\pi}^{\frac{1}{k}}$.

We will show that the time complexity of SIMULATED-PATH($d_0, t$) is independent of map resolution but the same procedure in \cite{LowAAMAS08} is not.
It is also less sensitive to increasing robot team size.
Assuming no prior data and $|\set{A}'(\ve{x}_i)| = \Delta$,
the time needed to evaluate the stagewise rewards $\mathbb{H}[Y_{x_{i+1}}|d_i]$ for all $\Delta$  new locations $x_{i+1}$ (i.e., using Cholesky factorization) is $\set{O}(t^3+\Delta t^2)$, which is independent of $|\set{X}|$ and results in $\set{O}(t(t^3+\Delta(t^2+\nu)))$ time to run SIMULATED-PATH($d_0, t$).
In contrast, the time needed to evaluate the stagewise rewards in \cite{LowAAMAS08} is $\set{O}(t^3 + \Delta(t^2+ |\set{X}|t)+|\set{X}|t^2)$, which depends on $|\set{X}|$ and entails  $\set{O}(t(t^3 + \Delta(t^2+ |\set{X}|t+\nu)+|\set{X}|t^2))$ time to run the same procedure.
When the joint action set size $\Delta$ increases with larger robot team size, the time to run the procedure in \cite{LowAAMAS08}
increases faster than that of ours due to the gradient factor $|\set{X}|t$ involving large domain size.
In the next section, we will report the time taken to run this procedure empirically.\vspace{-3mm}
\begin{algorithm}
\caption{URTDP ($\alpha$ is user-specified bound).}
\label{alg2}
\scriptsize{
URTDP($d_0, t$):
\begin{algorithmic}
\STATE {\bf while} $\overline{U}_0(d_0) - \underline{U}_0(d_0)> \alpha$ {\bf do} SIMULATED-PATH($d_0, t$)
%\WHILE{$\overline{U}_0(d_0) - \underline{U}_0(d_0)> \alpha$}
%\STATE SIMULATED-PATH($d_0, t$)
%\ENDWHILE
\STATE
\end{algorithmic}\vspace{-3mm}
SIMULATED-PATH($d_0, t$):\vspace{-1mm}
\begin{algorithmic}[1]
\STATE $i \leftarrow 0$
\WHILE{$i < t$}
\STATE $\ve{a}^{\ast}_i \leftarrow \arg\max_{\ve{a}_i} \overline{Q}_i$($\ve{a}_i, d_i$) \label{line:urtdp0}
\STATE $\forall j,$ $\Xi_j \leftarrow \underline{p}^{[ j]}_{x^{\ast}_{i+1}} \{
\overline{U}_{i+1}(d_i, x^{\ast}_{i+1}, \underline{z}^{[ j]}_{x^{\ast}_{i+1}}) - \underline{U}_{i+1}(d_i, x^{\ast}_{i+1}, \underline{z}^{[ j]}_{x^{\ast}_{i+1}})\}$\label{line:urtdp1}
\STATE $z \leftarrow$ sample from distribution at points $\underline{z}^{[ j]}_{x^{\ast}_{i+1}}$ of
probability $\Xi_j/\sum_k \Xi_k$ \label{line:urtdp4}
\STATE $d_{i+1} \leftarrow d_i, x^{\ast}_{i+1}, z$
\STATE $i \leftarrow i+1$
\ENDWHILE
\STATE $\overline{U}_{i}(d_i) \leftarrow \max_{\ve{a}_i} \mathbb{H}[Y_{x_{i+1}}|d_i], \ \underline{U}_{i}(d_i) \leftarrow \overline{U}_{i}(d_i)$
%\STATE $\underline{U}_{i}(d_i) \leftarrow \max_{\ve{a}_i} \mathbb{H}[Y_{x_{i+1}}|d_i]$
\WHILE{$i > 0$}
\STATE $i \leftarrow i - 1$
\STATE $\overline{U}_{i}$($d_i$) $\leftarrow \max_{\ve{a}_i} \overline{Q}_i$($\ve{a}_i, d_i$) \label{line:urtdp2}
\STATE $\underline{U}_{i}$($d_i$) $\leftarrow \max_{\ve{a}_i} \underline{Q}_i$($\ve{a}_i, d_i$) \label{line:urtdp3}
\ENDWHILE
\end{algorithmic}
}
\end{algorithm}\vspace{-4.5mm}
%\vspace{-3mm}
%
\section{Experiments and Discussion}
%\vspace{-1.5mm}
\label{sect:expt}
%\begin{figure*}
%\begin{tabular}{cccc}
%\hspace{-2mm}\multirow{3}{*}[1.7cm]{\epsfig{figure=planktongray,width=6cm}} & \hspace{-3mm}\epsfig{figure=mapabslgp,width=2.5cm} & 
%\multirow{3}{*}[1.7cm]{\epsfig{figure=kgray,width=6cm}} & \hspace{-3mm}\epsfig{figure=mapabslgpk,width=2.5cm}\vspace{-1mm}\\
%& \hspace{-3mm}(b) & & \hspace{-3mm}(e) \vspace{1mm}\\
%& \hspace{-3mm}\epsfig{figure=mapabsgp,width=2.5cm} & &  \hspace{-3mm}\epsfig{figure=mapabsgpk,width=2.5cm}\\
%\hspace{-2mm}(a) & \hspace{-3mm}(c) & \hspace{1mm}(d) & \hspace{-3mm}(f)\vspace{-4mm}
%\end{tabular}
%\caption{(a) Plankton density (chl-a) field with prediction error maps for (b) strictly adaptive $\underline{\pi}^{1/k}$ and (c) non-adaptive $\pi^n$:
%20 units (white circles) are randomly selected as prior data. The robots start at locations marked by `$\times$'s. The black and gray robot paths are produced by $\underline{\pi}^{1/k}$ and $\pi^n$ respectively. (d-f) Potassium (K) distribution field with error maps for $\underline{\pi}^{1/k}$ and $\pi^n$.
%Prediction error maps of chl-a (K) field for (b) ((e)) $\underline{\pi}^{1/k}$ and (c) ((f)) $\pi^n$.
%}
%\vspace{-4mm}
%\label{fig:pkmap}
%\end{figure*}
This section evaluates, empirically, the approximately optimal strictly adaptive policy $\underline{\pi}^{\frac{1}{k}}$ on $2$ real-world datasets exhibiting positive skew: (a) June 2006 plankton density data (Fig.~\ref{fig:pkmap}a) of Chesapeake Bay bounded within lat. $38.481-38.591$N and lon. $76.487-76.335$W, and (b) potassium distribution data (Fig.~\ref{fig:pkmap}d) of Broom's Barn farm spanning $520$m by $440$m. 
Each region is discretized into a $14\times12$ grid of sampling units. 
Each unit $x$ is, respectively, associated with (a) plankton density $y_{x}$ (chl-a) in mg~$\mbox{m}^{-3}$, and (b) potassium level $y_{x}$ (K) in mg~$\mbox{l}^{-1}$. Each region comprises, respectively, (a) $|\set{X}|=148$ and (b) $|\set{X}|=156$ such units.
Using a team of $2$ robots, each robot is tasked to explore 9 adjacent units in its path including its starting unit.
If only $1$ robot is used, it is placed, respectively, in (a) top and (b) bottom starting unit, and samples all 18 units.
Each robot's actions are restricted to move to the front, left, or right unit.
We use the data of 20 randomly selected units to learn the hyperparameters (i.e., mean and covariance structure) of GP and $\ell$GP through maximum likelihood estimation \cite{Rasmussen06}.
So, prior data $d_0$ comprise the randomly selected and robot starting units.
\begin{figure}
\begin{tabular}{cccc}
\multicolumn{2}{c}{\hspace{-9mm}\epsfig{figure=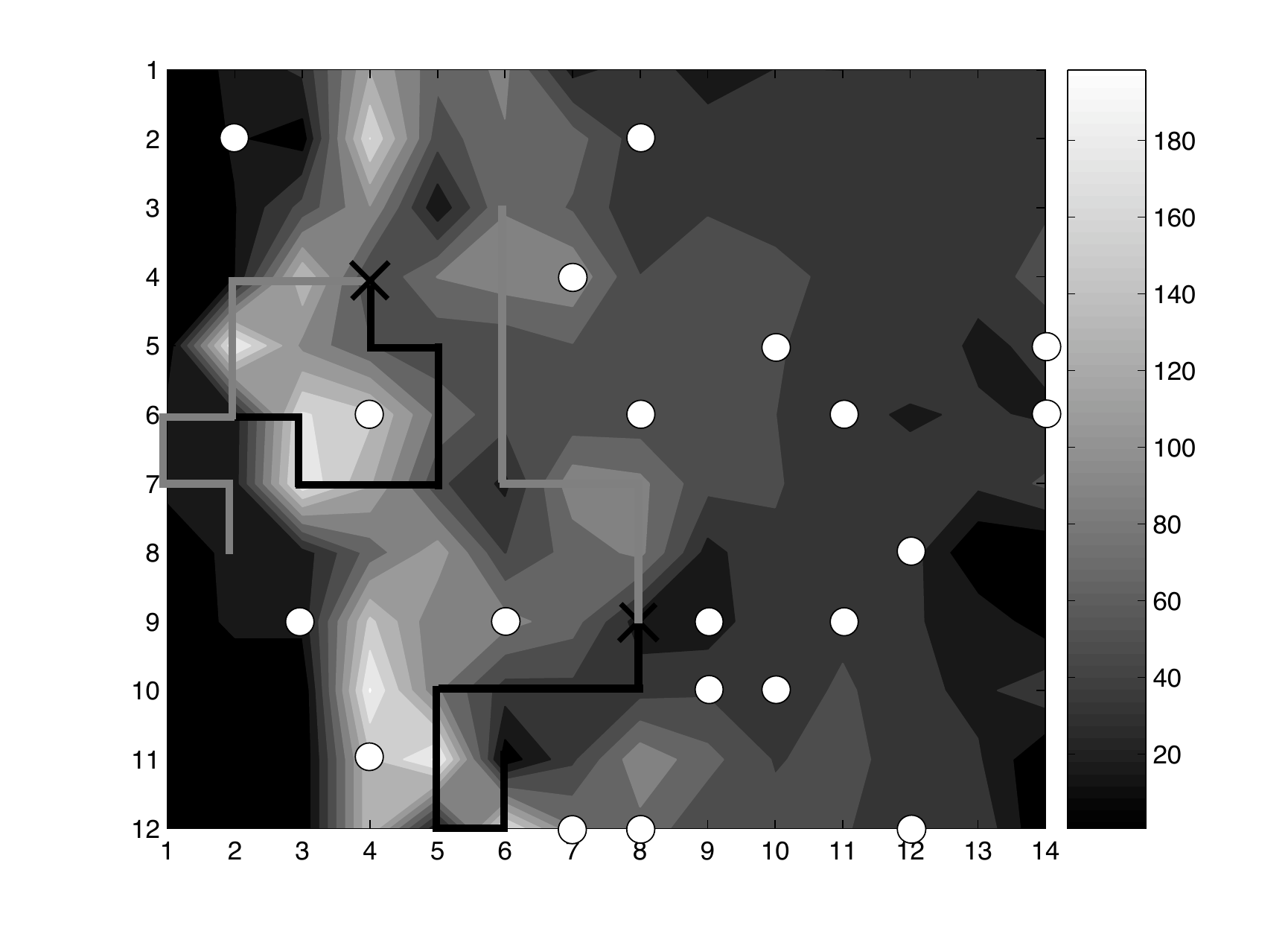,width=4.7cm}} & \multicolumn{2}{c}{\hspace{-3mm}\epsfig{figure=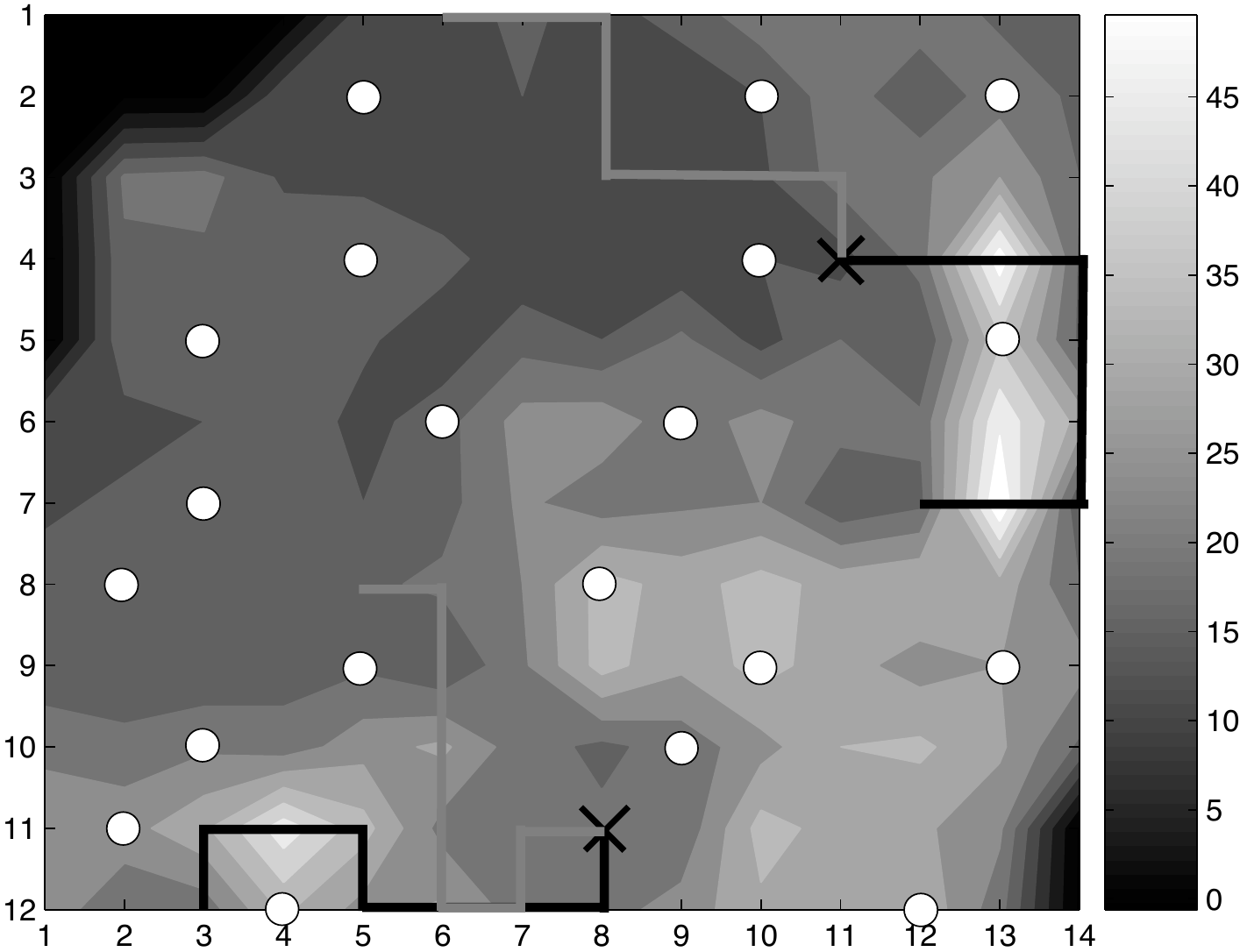,width=4.7cm}}\vspace{-1.5mm}\\
\multicolumn{2}{c}{\hspace{-7mm}\footnotesize(a)} & \multicolumn{2}{c}{\hspace{-4mm}\footnotesize(d)}\\
\hspace{-9mm}\epsfig{figure=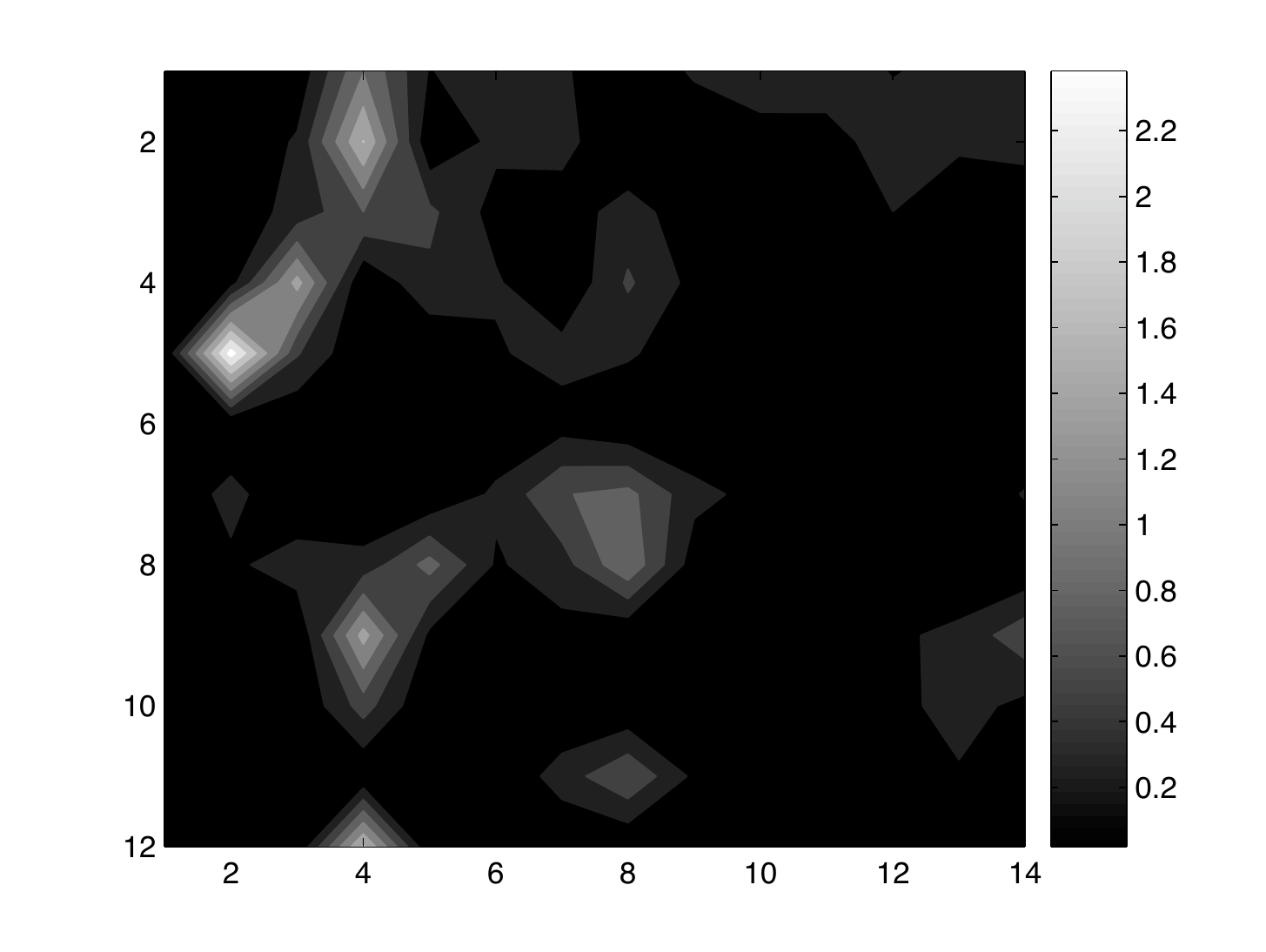,width=2.3cm} & \hspace{-4mm}\epsfig{figure=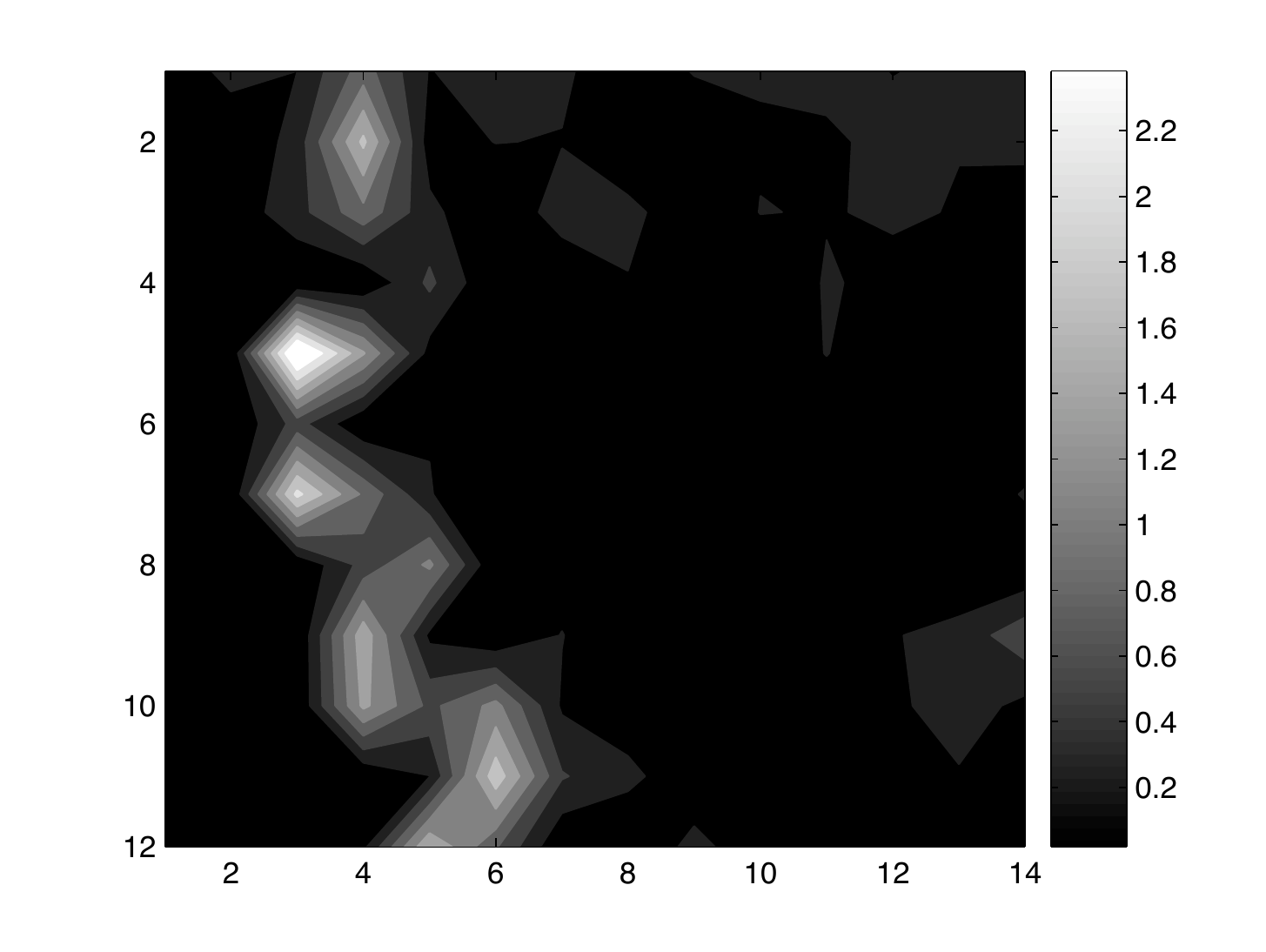,width=2.3cm} & \hspace{-3mm}\epsfig{figure=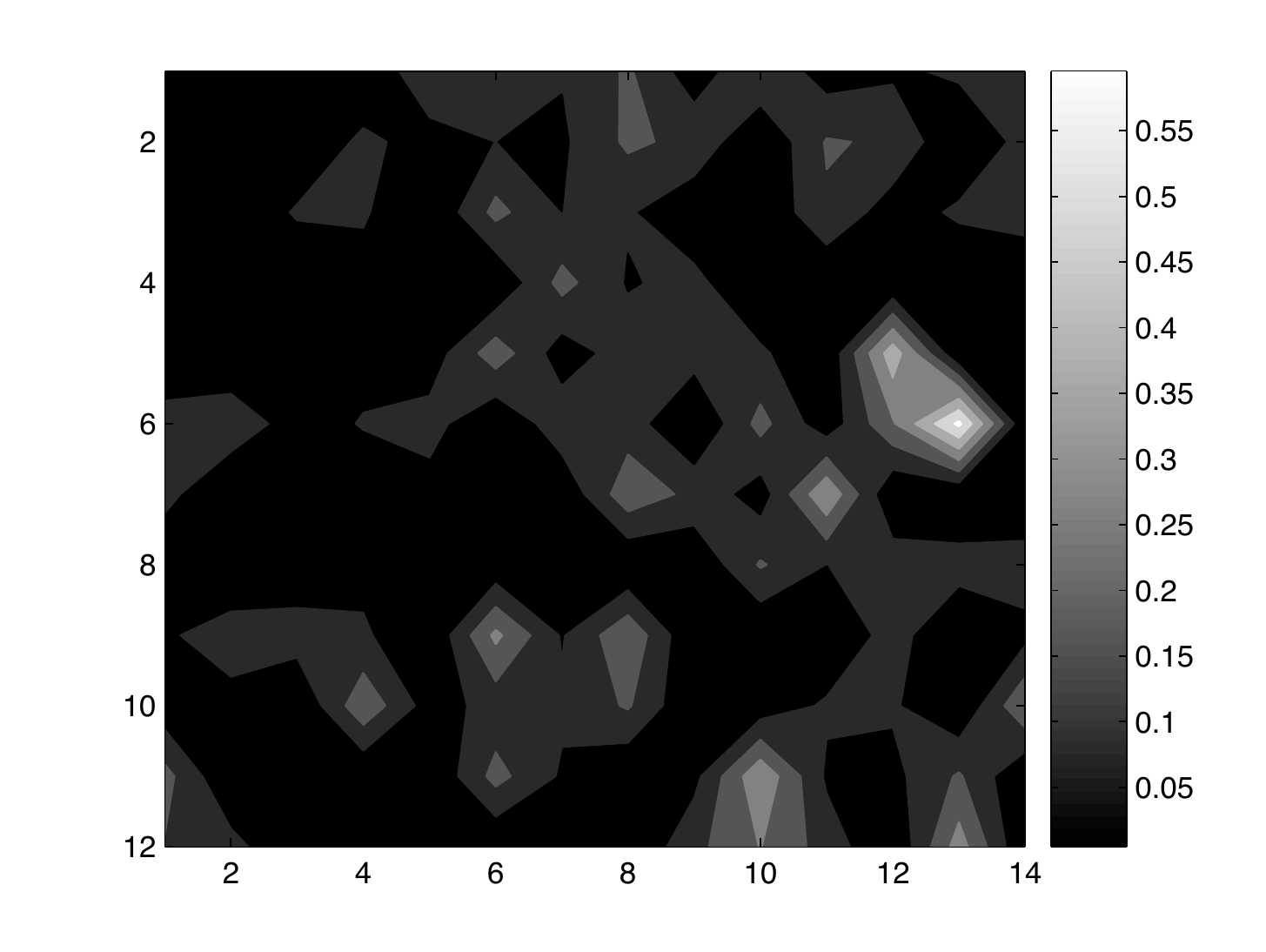,width=2.3cm} & \hspace{-4mm}\epsfig{figure=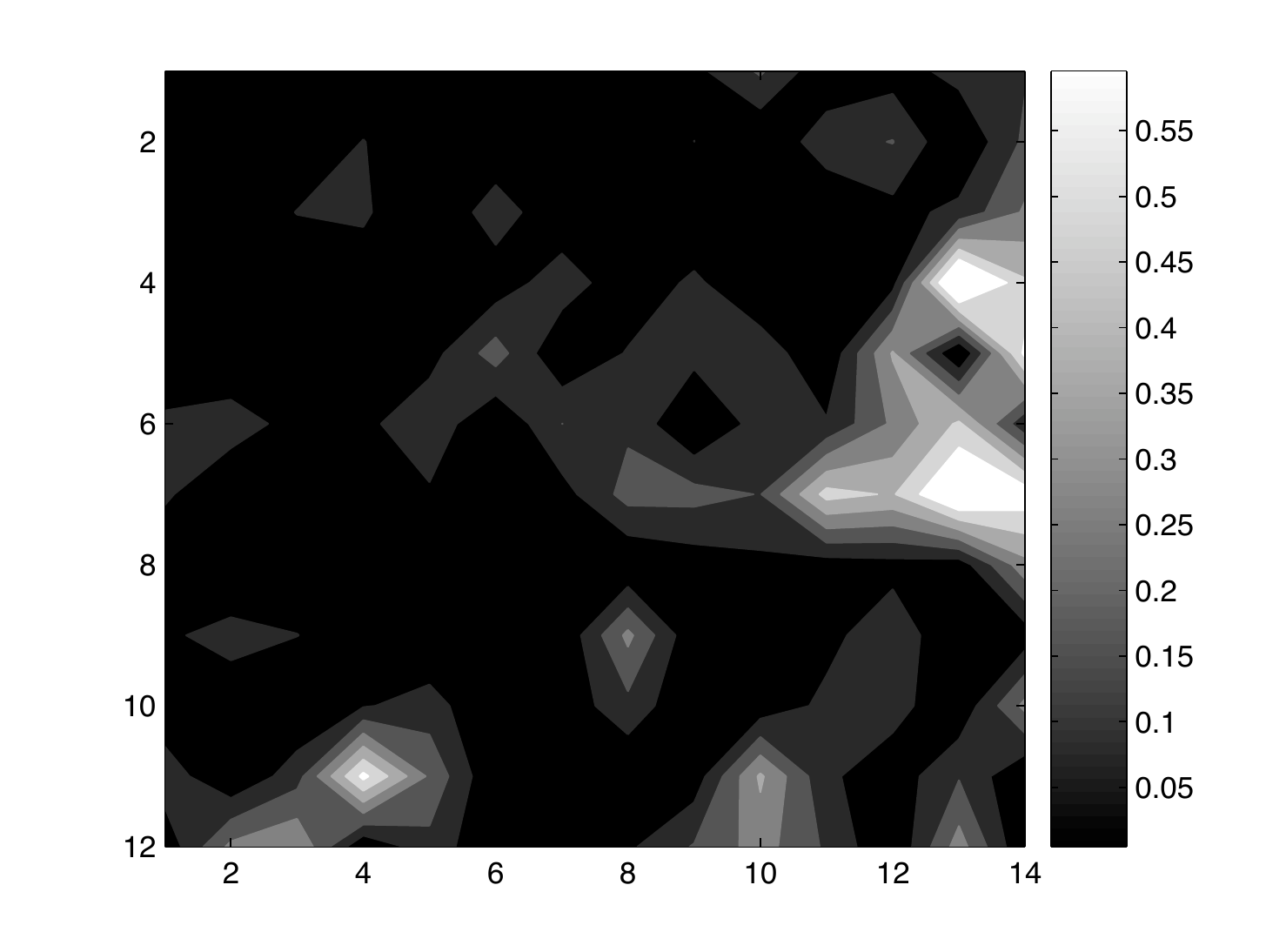,width=2.3cm}\vspace{-1.5mm}\\
\hspace{-9mm}{\footnotesize(b)} & \hspace{-4mm}{\footnotesize(c)} & \hspace{-3mm}{\footnotesize(e)} & \hspace{-4mm}{\footnotesize(f)}\vspace{-4mm}
\end{tabular}
\caption{(a) chl-a field with prediction error maps for (b) strictly adaptive $\underline{\pi}^{1/k}$ and (c) non-adaptive $\pi^n$:
20 units (white circles) are randomly
selected as prior data. The robots start at locations marked by `$\times$'s. The black and gray
robot paths are produced by $\underline{\pi}^{1/k}$
and $\pi^n$ respectively. (d-f) K field with error maps for $\underline{\pi}^{1/k}$
and $\pi^n$.
%Prediction error maps of chl-a (K) field for (b) ((e)) $\underline{\pi}^{1/k}$ and (c) ((f)) $\pi^n$.
}
\vspace{-6mm}
\label{fig:pkmap}
\end{figure}

The performance of $\underline{\pi}^{\frac{1}{k}}$ is compared to the policies produced by four state-of-the-art exploration strategies:
The \emph{optimal non-adaptive policy $\pi^n$ for GP} \cite{Shewry87} is produced by solving $i$MASP($n$) (\ref{eq:3r}). Similar to Theorem~\ref{thm:2}, it can be shown to be equivalent to the strictly adaptive $\pi^{\frac{1}{k}}$ for GP.
Although $\underline{i}$MASP($\frac{1}{k}$) and $i$MASP($n$) can be solved exactly, their state size grows exponentially with the number of stages. To alleviate this computational difficulty, 
we use anytime heuristic search algorithms URTDP (Algorithm~\ref{alg2}) and Learning Real-Time A$^{\ast}$ 
%\cite{Korf90} 
to, respectively, solve $\underline{i}$MASP($\frac{1}{k}$) and $i$MASP($n$) approximately.
The \emph{adaptive greedy policy for $\ell$GP} repeatedly chooses a reward-maximizing action (i.e., by repeatedly solving $i$MASP($\frac{1}{k}$) with $t = 0$ in (\ref{eq:4r4})) to form the paths. The \emph{non-adaptive greedy policy for GP} performs likewise but does it in the log-scale.
In contrast to the above policies that optimize the entropy criterion (\ref{eq:1a}), a non-adaptive greedy policy is proposed by \citeauthor{Guestrin05} \shortcite{Guestrin05} to approximately maximize the mutual information (MI) criterion for GP; it repeatedly selects a new sampling location that maximizes the increase in MI. We call this the \emph{MI-based policy}.
%It is not clear how this greedy policy can be revised to be adaptive for use in $\ell$GP as it does not seem possible to evaluate the increase in conditional MI.
\vspace{1mm}\\
{\bf Performance metrics}. 
Two metrics are used to evaluate the above policies: (a) \emph{Posterior map \underline{ent}ropy} (ENT) $\mathbb{H}[\ve{Y}_{\overline{\ve{x}}_{0:t}}|d_t]$ of domain $\set{X}$ is the optimized criterion (\ref{eq:1a}) measuring the posterior joint entropy of the original measurements $\ve{Y}_{\overline{\ve{x}}_{0:t}}$ at the unobserved locations $\overline{\ve{x}}_{0:t}$ where $t = 16$ ($17$) for the case of 2 (1) robots. A smaller ENT implies lower map uncertainty;
(b) \emph{Mean-squared relative \underline{err}or} (ERR) $|\set{X}|^{-1} \sum_{x\in \set{X}} \{(y_{x} - \mu_{Y_{x}|d_{t}})/\bar{\mu}\}^2$ measures the posterior map error from using the best unbiased predictor $\mu_{Y_{x}|d_{t}}$ (i.e., $\ell$GP posterior mean) \cite{LowAAMAS08} of the measurement $y_x$ to predict the hotspot field where $\bar{\mu} = |\set{X}|^{-1} \sum_{x\in \set{X}} y_{x}$. 
Although this criterion is not the one being optimized, it allows the use of ground truth measurements to evaluate if the field is being mapped accurately.
A smaller ERR implies lower map prediction error.\vspace{1mm}

Table~\ref{tab:compare} shows the results of various policies with different assumed models and robot team sizes for chl-a and K fields. For  $\underline{i}$MASP($\frac{1}{k}$) and $i$MASP($n$), the results are obtained using the policies provided by the anytime algorithms after running $120000$ simulated paths.
The differences in results between policies have been verified using $t$-tests ($\alpha=0.1$) to be statistically significant.
\vspace{1mm}\\
{\bf Plankton density data}. 
The results show that
%policies for $\ell$GP achieve lower ENT and ERR than that of GP.
the strictly adaptive $\underline{\pi}^{\frac{1}{k}}$ achieves lowest ENT and ERR as compared to the tested policies. From Fig.~\ref{fig:pkmap}a, $\underline{\pi}^{\frac{1}{k}}$ moves the robots to sample the hotspots showing higher spatial variability whereas $\pi^n$ moves them to sparsely sampled areas.
Figs.~\ref{fig:pkmap}b and~\ref{fig:pkmap}c show, respectively, the prediction error maps resulting from $\underline{\pi}^{\frac{1}{k}}$ and $\pi^n$; the prediction error at each location $x$ is measured using $|y_{x} - \mu_{Y_{x}|d_{t}}|/\bar{\mu}$. Locations with large errors are mostly concentrated in the left region where the field is highly-varying and contains higher measurements. Compared to $\underline{\pi}^{\frac{1}{k}}$, $\pi^n$ incurs large errors at more locations in or close to hotspots, thus resulting in higher ERR.

We also compare the time needed to run the first $10000$ SIMULATED-PATH($d_0, t$)'s of our URTDP algorithm to that of \citeauthor{LowAAMAS08} \shortcite{LowAAMAS08}, which are $115$s and $10340$s respectively for $2$ robots  (i.e., $90\times$ faster).
They, respectively, take $66$s and $2835$s for $1$ robot (i.e., $43\times$ faster).
So, scaling to $2$ robots incurs $1.73\times$ and $3.65\times$ more time for the respective algorithms.
Policy $\underline{\pi}^{\frac{1}{k}}$ can already achieve the performance reported in Table~\ref{tab:compare} for $2$ robots, and ENT of $389.23$ and ERR of $0.231$ for $1$ robot. In contrast, the policy of \citeauthor{LowAAMAS08} \shortcite{LowAAMAS08} only improves to ENT of $377.82$ ($391.85$) and ERR of $0.233$ ($0.252$) for $2$ ($1$) robots, which are slightly worse off.
% 1 robot: 2834.565589s (0.211529 389.178711, 0.251746 391.850800) 66.417101s (0.230613 389.233337)
%10000 simulated paths 10340.658s 114.725s
\vspace{1mm}\\
{\bf Potassium distribution data}.
The results show again that $\underline{\pi}^{\frac{1}{k}}$ achieves lowest ENT and ERR. From Fig.~\ref{fig:pkmap}d, $\underline{\pi}^{\frac{1}{k}}$ again moves the robots to sample the hotspots showing higher spatial variability whereas $\pi^n$ moves them to sparsely sampled areas.
Compared to $\underline{\pi}^{\frac{1}{k}}$, $\pi^n$ incurs large errors at a greater number of locations in or close to hotspots as shown in Figs.~\ref{fig:pkmap}e and~\ref{fig:pkmap}f, thus resulting in higher ERR.
\begin{table}
\vspace{-2.5mm}
\caption{Performance comparison of information-theoretic policies for chl-a and K fields: 1R (2R) denotes 1 (2) robots.}
\label{tab:compare}
\centering
\scriptsize{
\begin{tabular}{|l|c|c|c|c|c|}
\hline
\multicolumn{2}{|l|}{Plankton density (chl-a) field} & \multicolumn{2}{|c|}{ENT} & \multicolumn{2}{|c|}{ERR} \\
\hline
Exploration policy & Model & 1R & 2R & 1R & 2R \\
\hline
Adaptive $\underline{\pi}^{1/k}$ & $\ell$GP & 381.37 & 376.19 & 0.1827 & 0.2319 \\
%Non-adaptive $\pi^n$ & $\ell$GP & 0.277 & 0.279 & 1553 & 1522 \\
Adaptive greedy & $\ell$GP & 382.97 & 383.55 & 0.2919 & 0.2579 \\
Non-adaptive $\pi^n$ & GP & 390.62 & 399.63 & 0.4145 & 0.3194 \\
Non-adaptive greedy & GP & 392.35 & 392.51 & 0.2994 & 0.3356 \\
MI-based & GP & 395.37 & 397.02 & 0.2764 & 0.2706 \\
\hline
\end{tabular}
\begin{tabular}{|l|c|c|c|c|c|}
\hline
\multicolumn{2}{|l|}{Potassium (K) field} & \multicolumn{2}{|c|}{ENT} & \multicolumn{2}{|c|}{ERR} \\
\hline
Exploration policy & Model & 1R & 2R & 1R & 2R \\
\hline
Adaptive $\underline{\pi}^{1/k}$ & $\ell$GP & 47.330 & 48.287 & 0.0299 & 0.0213 \\
Adaptive greedy & $\ell$GP & 61.080 & 56.181 & 0.0457 & 0.0302 \\
Non-adaptive $\pi^n$ & GP & 67.084 & 59.318 & 0.0434 & 0.0358 \\
Non-adaptive greedy & GP & 58.704 & 64.186 & 0.0431 & 0.0335 \\
MI-based & GP & 59.058 & 67.390 & 0.0435 & 0.0343 \\
\hline
\end{tabular}}\vspace{-4mm}
\end{table}

To run $10000$ SIMULATED-PATH($d_0, t$)'s, our URTDP algorithm is $84\times$ ($48\times$) faster than that of \citeauthor{LowAAMAS08} \shortcite{LowAAMAS08} for $2$ ($1$) robots.
%exhibits similar computational gain (i.e., $84\times$ ($48\times$) faster for $2$ ($1$) robots).
% as that described in the previous subsection.
Scaling to $2$ robots incurs $1.93\times$ and $3.37\times$ more time for the respective algorithms.
%our URTDP and Low \emph{et al.}~\shortcite{LowAAMAS08}'s algorithms respectively.
Policy $\underline{\pi}^{\frac{1}{k}}$ can already achieve the performance reported in Table~\ref{tab:compare} for $1$ and $2$ robots.
In contrast, the policy of \citeauthor{LowAAMAS08} \shortcite{LowAAMAS08} achieves worse ENT of $67.132$ ($55.015$) for $2$ ($1$) robots. It achieves worse ERR of $0.032$ for $2$ robots but better ERR of $0.025$ for $1$ robot.
%\vspace{1.1mm}
% 1 robot: 2945.598691s (0.024923 55.015251) 61.394151s (same as above)
% 2 robots: 10000 simulated paths 9921.456s 118.762s
\vspace{1mm}\\
{\bf Summary of test results}. The above results show that the strictly adaptive $\underline{\pi}^{\frac{1}{k}}$ can learn the highest-quality hotspot field map (i.e., lowest ENT and ERR) among the tested state-of-the-art strategies.
After evaluating whether MASP- vs. $i$MASP-based planners are time-efficient for real-time deployment, we observe that $\underline{\pi}^{\frac{1}{k}}$ can achieve mapping performance comparable to the policy of \citeauthor{LowAAMAS08} \shortcite{LowAAMAS08} using significantly less time, and the incurred planning time is also less sensitive to larger robot team size.
Lastly, we see in Fig.~\ref{fig:pkmap} that the strictly adaptive $\underline{\pi}^{\frac{1}{k}}$ has exploited clustering phenomena (i.e., hotspots) to
achieve lower ENT and ERR than that of the non-adaptive $\pi^n$.
\vspace{-1mm}
\section{Conclusion}
%\vspace{-1mm}
% say something about insufficiency of GP to represent hotspot fields, which occur often in nature.
%
This paper describes an information-theoretic approach to efficient adaptive path planning
for active exploration and mapping of hotspot fields. We have shown that, like MASP, $i$MASP is capable of exploiting clustering phenomena to produce lower map uncertainty.
In contrast to MASP, the time complexity of solving (reward-maximizing) $i$MASP approximately is independent of map resolution and is also less sensitive to increasing robot team size as demonstrated theoretically and empirically.
This is clearly advantageous in large-scale, high-resolution exploration and mapping.
The proposed approximation techniques can be generalized to solve $i$MASPs that utilize the full joint action space of the robot team, thus allowing the robots to move simultaneously at every stage and the mission time to be constrained.
%For our future work, we will integrate the $i$MASP-based planner with a reactive controller to navigate the robots along the policy-planned paths for sampling.  
%\vspace{-6mm}
%The anytime algorithm URTDP can also be decentralized in a way that is described in \cite{Shoham08} so that every robot in the team can execute it simultaneously, thus improving the algorithm's convergence performance.
\vspace{1mm}\\
{\bf Acknowledgments.}
%\vspace{-1mm}
We would like to thank Dr R. Webster from Rothamsted Research for providing the Broom's Barn Farm data.
%\vspace{-6mm}

\begin{footnotesize}
\bibliography{adaptivesampling}
\bibliographystyle{aaai}
\end{footnotesize}

\appendix
\section{Proofs}
\subsection{Theorem~\ref{thm:1}}
\label{sect:app1}
\emph{Proof by induction} on $i$ that
$\displaystyle 
V^{\pi^1}_i(d_{i}) =
\displaystyle\mathbb{H}[\ve{Z}_{\overline{\ve{x}}_{0:i}}|d_i] - U^{\pi^1}_{i}(d_{i})$ for $i = n-1, \ldots, 0$.\\

\noindent
\emph{Base case} ($i = n-1$):
$$
\begin{array}{l}
\displaystyle V^{\pi^1}_{n-1}(d_{n-1}) \\
= \displaystyle\min_{\ve{a}_{n-1} \in \set{A}(\ve{x}_{n-1})} 
\int f(\ve{z}_{\ve{x}_{n}}| d_{n-1}) \ V^{\pi^1}_{n}(d_{n}) \ \mbox{d}\ve{z}_{\ve{x}_{n}} \\
= \displaystyle\min_{\ve{a}_{n-1} \in \set{A}(\ve{x}_{n-1})} 
\int f(\ve{z}_{\ve{x}_{n}}| d_{n-1}) \ \mathbb{H}[\ve{Z}_{\overline{\ve{x}}_{0:n}}| d_n] \ \mbox{d}\ve{z}_{\ve{x}_{n}} \\
= \displaystyle\min_{\ve{a}_{n-1} \in \set{A}(\ve{x}_{n-1})} \mathbb{H}[\ve{Z}_{\ve{x}_{n}}, \ve{Z}_{\overline{\ve{x}}_{0:n}}\mid d_{n-1}] - \mathbb{H}[\ve{Z}_{\ve{x}_{n}} | d_{n-1} ]\\
= \displaystyle\min_{\ve{a}_{n-1} \in \set{A}(\ve{x}_{n-1})} \mathbb{H}[\ve{Z}_{\overline{\ve{x}}_{0:n-1}}| d_{n-1}] - \mathbb{H}[\ve{Z}_{\ve{x}_{n}} | d_{n-1} ]\\
= \displaystyle \mathbb{H}[\ve{Z}_{\overline{\ve{x}}_{0:n-1}}| d_{n-1}] - \max_{\ve{a}_{n-1} \in \set{A}(\ve{x}_{n-1})}\mathbb{H}[\ve{Z}_{\ve{x}_{n}} | d_{n-1} ]\\
= \displaystyle \mathbb{H}[\ve{Z}_{\overline{\ve{x}}_{0:n-1}}| d_{n-1}] - U^{\pi^1}_{n-1}(d_{n-1}) \ .
\end{array}$$
The first and second equalities follow from (\ref{eq:4}).
The third equality is due to the chain rule for entropy (Cover and Thomas 1991).
%\cite{Cover91}.
The last equality is due to (\ref{eq:4r}).
Hence, the base case is true.\\

\noindent
\emph{Inductive case}: Suppose that 
\begin{equation}
\displaystyle 
\displaystyle 
V^{\pi^1}_i(d_{i}) =\mathbb{H}[\ve{Z}_{\overline{\ve{x}}_{0:i}}|d_i] - U^{\pi^1}_{i}(d_{i})
\label{eq:a1}
\end{equation}
is true. We have to prove that 
$V^{\pi^1}_{i-1}(d_{i-1}) =\mathbb{H}[\ve{Z}_{\overline{\ve{x}}_{0:i-1}}| d_{i-1}] - U^{\pi^1}_{i-1}(d_{i-1})$ is true.
$$
\hspace{-2mm}
\begin{array}{l}
\displaystyle V^{\pi^1}_{i-1}(d_{i-1}) \\
= \displaystyle\min_{\ve{a}_{i-1} \in \set{A}(\ve{x}_{i-1})} 
\int f(\ve{z}_{\ve{x}_{i}}|d_{i-1}) \ V^{\pi^1}_{i}(d_{i}) \ \mbox{d}\ve{z}_{\ve{x}_{i}} \\
= \displaystyle\min_{\ve{a}_{i-1} \in \set{A}(\ve{x}_{i-1})} 
\int f(\ve{z}_{\ve{x}_{i}}| d_{i-1}) \left(\mathbb{H}[\ve{Z}_{\overline{\ve{x}}_{0:i}}| d_i] - U^{\pi^1}_{i}(d_{i})\right) \mbox{d}\ve{z}_{\ve{x}_{i}} \\
= \displaystyle\min_{\ve{a}_{i-1} \in \set{A}(\ve{x}_{i-1})} 
\mathbb{H}[\ve{Z}_{\overline{\ve{x}}_{0:i-1}}| d_{i-1}] - \mathbb{H}[\ve{Z}_{\ve{x}_{i}} | d_{i-1} ] \ -\\
\ \ \ \ \ \ \ \ \ \ \ \ \ \ \ \ \ \ \ \ \ \ \ \ \ \displaystyle \int f(\ve{z}_{\ve{x}_{i}}| d_{i-1}) \ U^{\pi^1}_{i}(d_{i}) \ \mbox{d}\ve{z}_{\ve{x}_{i}} \\
= \displaystyle\mathbb{H}[\ve{Z}_{\overline{\ve{x}}_{0:i-1}}| d_{i-1}]-\max_{\ve{a}_{i-1} \in \set{A}(\ve{x}_{i-1})}\Big(\mathbb{H}[\ve{Z}_{\ve{x}_{i}} | d_{i-1} ] \ + \\
\ \ \ \ \displaystyle\int f(\ve{z}_{\ve{x}_{i}}| d_{i-1}) \ U^{\pi^1}_{i}(d_{i}) \ \mbox{d}\ve{z}_{\ve{x}_{i}}\Big) \\
= \displaystyle\mathbb{H}[\ve{Z}_{\overline{\ve{x}}_{0:i-1}}| d_{i-1}]- U^{\pi^1}_{i-1}(d_{i-1}) \ .
\end{array}
$$
The first equality follows from (\ref{eq:4}). The second equality follows from (\ref{eq:a1}). The third equality follows from linearity of expectation and the chain rule for entropy (Cover and Thomas 1991). 
%\cite{Cover91}
The last equality is due to (\ref{eq:4r}).
Hence, the inductive case is true. 

It is clear from above that the induced optimal adaptive policies from solving the cost-minimizing and reward-maximizing $i$MASP($1$)'s coincide.
\subsection{Equation~\ref{eq:7a}}
\label{sect:app2}
Since
$f(\ve{Z}_{\ve{x}_{i+1}} = \ve{z}_{\ve{x}_{i+1}}| d_{i}) =$ $$\displaystyle\frac{\exp\left\{-\displaystyle\frac{1}{2} (\ve{z}_{\ve{x}_{i+1}}-\mu_{\ve{Z}_{\ve{x}_{i+1}}\mid d_{i}}) \Sigma_{\ve{Z}_{\ve{x}_{i+1}}\mid d_{i}}^{-1} (\ve{z}_{\ve{x}_{i+1}}-\mu_{\ve{Z}_{\ve{x}_{i+1}}\mid d_{i}})^{\top} \right\}}{\sqrt{(2\pi)^k \ |\Sigma_{\ve{Z}_{\ve{x}_{i+1}}\mid d_{i}}|}} ,$$
$$
\hspace{-2mm}
\begin{array}{l}
\mathbb{H}[\ve{Z}_{\ve{x}_{i+1}}| d_i] \vspace{1mm}\\
= \mathbb{E}[- \log f(\ve{Z}_{\ve{x}_{i+1}}| d_i) | d_{i}] \vspace{1mm}\\
= \displaystyle\mathbb{E}[ \log \sqrt{(2\pi)^k |\Sigma_{\ve{Z}_{\ve{x}_{i+1}}\mid d_i}|} \ +\\
\ \ \ \ \displaystyle\frac{1}{2} (\ve{Z}_{\ve{x}_{i+1}}-\mu_{\ve{Z}_{\ve{x}_{i+1}}\mid d_{i}}) \Sigma_{\ve{Z}_{\ve{x}_{i+1}}\mid d_{i}}^{-1} (\ve{Z}_{\ve{x}_{i+1}}-\mu_{\ve{Z}_{\ve{x}_{i+1}}\mid d_{i}})^{\top} | d_{i}] \\
= \displaystyle\log \sqrt{(2\pi)^k |\Sigma_{\ve{Z}_{\ve{x}_{i+1}}\mid d_i}|} \ +\\
\ \ \ \ \displaystyle\frac{1}{2} \mathbb{E}[(\ve{Z}_{\ve{x}_{i+1}}-\mu_{\ve{Z}_{\ve{x}_{i+1}}\mid d_{i}}) \Sigma_{\ve{Z}_{\ve{x}_{i+1}}\mid d_{i}}^{-1} (\ve{Z}_{\ve{x}_{i+1}}-\mu_{\ve{Z}_{\ve{x}_{i+1}}\mid d_{i}})^{\top} | d_{i}]\vspace{1mm}\\
= \displaystyle\log \sqrt{(2\pi)^k |\Sigma_{\ve{Z}_{\ve{x}_{i+1}}\mid d_i}|} \ +\\
\ \ \ \ \displaystyle\frac{1}{2} \mathbb{E}[\mbox{tr}(\Sigma_{\ve{Z}_{\ve{x}_{i+1}}\mid d_{i}}^{-1} (\ve{Z}_{\ve{x}_{i+1}}-\mu_{\ve{Z}_{\ve{x}_{i+1}}\mid d_{i}}) (\ve{Z}_{\ve{x}_{i+1}}-\mu_{\ve{Z}_{\ve{x}_{i+1}}\mid d_{i}}))| d_i]\vspace{1mm}\\
= \displaystyle\log \sqrt{(2\pi)^k |\Sigma_{\ve{Z}_{\ve{x}_{i+1}}\mid d_i}|} \ +\\
\ \ \ \ \displaystyle\frac{1}{2} \mbox{tr}(\Sigma_{\ve{Z}_{\ve{x}_{i+1}}\mid d_{i}}^{-1} \mathbb{E}[(\ve{Z}_{\ve{x}_{i+1}}-\mu_{\ve{Z}_{\ve{x}_{i+1}}\mid d_{i}}) (\ve{Z}_{\ve{x}_{i+1}}-\mu_{\ve{Z}_{\ve{x}_{i+1}}\mid d_{i}})| d_i])\vspace{1mm}\\
= \displaystyle\log \sqrt{(2\pi)^k |\Sigma_{\ve{Z}_{\ve{x}_{i+1}}\mid d_i}|} + \displaystyle\frac{1}{2} \mbox{tr}(\Sigma_{\ve{Z}_{\ve{x}_{i+1}}\mid d_{i}}^{-1} \Sigma_{\ve{Z}_{\ve{x}_{i+1}}\mid d_{i}})\vspace{1mm}\\
= \displaystyle\log \sqrt{(2\pi)^k |\Sigma_{\ve{Z}_{\ve{x}_{i+1}}\mid d_i}|} + \frac{1}{2} \mbox{tr}(\ma{I})\vspace{1mm}\\
= \displaystyle\log \sqrt{(2\pi)^k |\Sigma_{\ve{Z}_{\ve{x}_{i+1}}\mid d_i}|} + \frac{k}{2}\\
= \displaystyle\log \sqrt{(2\pi e)^k |\Sigma_{\ve{Z}_{\ve{x}_{i+1}}\mid d_i}|} \ .
\end{array}
$$
The fourth equality is due to the trace property tr($AB$) = tr($BA$).
\subsection{Equation~\ref{eq:11a}}
\label{sect:app3}
Using the Jacobian method of variable transformation,
$$
\hspace{-2mm}
\begin{array}{rl}
f(\ve{Y}_{\ve{x}_{i+1}}| d_{i}) =& f(\ve{Z}_{\ve{x}_{i+1}}| d_{i}) \displaystyle\prod_{x\in\set{X}'}\frac{d Z_{x}}{d Y_{x}}\\
=& f(\ve{Z}_{\ve{x}_{i+1}}| d_{i}) \displaystyle\prod_{x\in\set{X}'} \frac{1}{Y_{x}}
\end{array}
$$
where $\set{X}' = \{x\mid x \ \mbox{is a location component in} \ \ve{x}_{i+1} \}$.
So,
$$
\begin{array}{l}
\mathbb{H}[\ve{Y}_{\ve{x}_{i+1}}| d_{i}]\vspace{1mm}\\
= \mathbb{E}[- \log f(\ve{Y}_{\ve{x}_{i+1}}| d_{i}) | d_{i}] \\
= \mathbb{E}[- \log \displaystyle\left( f(\ve{Z}_{\ve{x}_{i+1}}| d_{i}) \displaystyle\prod_{x\in\set{X}'} \frac{1}{Y_{x}}\right) | d_{i}] \\
= \mathbb{E}[- \log f(\ve{Z}_{\ve{x}_{i+1}}| d_{i}) + \displaystyle\sum_{x\in\set{X}'} \log Y_{x} | d_{i}]\\
= \mathbb{E}[- \log f(\ve{Z}_{\ve{x}_{i+1}}| d_{i}) | d_{i}] + \displaystyle\sum_{x\in\set{X}'} \mathbb{E}[Z_{x} | d_{i}]\\
= \displaystyle \log \sqrt{(2\pi e)^k |\Sigma_{\ve{Z}_{\ve{x}_{i+1}}\mid d_{i}}|} + \bm{\mu}_{\ve{Z}_{\ve{x}_{i+1}}|d_i} \ve{1}^{\top} \ .
\end{array}
$$
The fourth equality is due to the transformation $Z_{x} = \log Y_{x}$ and linearity of expectation.
The fifth equality follows from (\ref{eq:7a}).
\subsection{Lemma~\ref{lem:3}}
\label{sect:app4}
We first show that $\mathbb{H}[Y_{x_{i+1}}| d_i]$ is convex in $\ve{z}_{\ve{x}_{0:i}}$ for $i = 0,\ldots,t$.
From (\ref{eq:11a}), we know that 
$$\mathbb{H}[Y_{x_{i+1}}| d_i] = \log\sqrt{2\pi e\sigma^2_{Z_{x_{i+1}}|d_i}} + \mu_{Z_{x_{i+1}}|d_i} \ .$$
From (\ref{eq:6}),
the posterior mean $\mu_{Z_{x_{i+1}}|d_i}$ is an affine function of $\ve{z}_{\ve{x}_{0:i}}$. Hence, it is convex in $\ve{z}_{\ve{x}_{0:i}}$ ((Boyd and Vandenberghe 2004), pp. 71). 
%\cite{Boyd04}
From (\ref{eq:7}), the posterior variance $\sigma^2_{Z_{x_{i+1}}|d_i}$ is independent of $\ve{z}_{\ve{x}_{0:i}}$. So, $\log\sqrt{2\pi e\sigma^2_{Z_{x_{i+1}}|d_i}}$ is a constant term.
Therefore, $\mathbb{H}[Y_{x_{i+1}}| d_i]$ is convex in $\ve{z}_{\ve{x}_{0:i}}$.

We will revert to using $Z_{x_{i+1}}$ in $i$MASP($\frac{1}{k}$) (\ref{eq:4r4}) for $\ell$GP (i.e., by transforming $Z_{x_{i+1}}=\log Y_{x_{i+1}}$).\\

\noindent
\emph{Proof by induction} on $i$ that $U_{i}(d_{i})$ is convex in
$\ve{z}_{\ve{x}_{0:i}}$ for $i = t, \ldots, 0$.\\
\\
\noindent
\emph{Base case} ($i = t$): As proven above, $\mathbb{H}[Y_{x_{t+1}}| d_t]$ is convex in $\ve{z}_{\ve{x}_{0:t}}$. Then, the pointwise maximum of $\mathbb{H}[Y_{x_{t+1}}| d_t]$ (i.e., $\max_{\ve{a}_{t} \in \set{A}'(\ve{x}_t)}
\mathbb{H}[Y_{x_{t+1}}| d_t]$) is convex in $\ve{z}_{\ve{x}_{0:t}}$ ((Boyd and Vandenberghe 2004), pp. 81). Therefore, $U_{t}(d_{t})$ is convex in $\ve{z}_{\ve{x}_{0:t}}$. The base case is true.\\
\\%\cite{Boyd04}
\noindent
\emph{Inductive case}: Suppose that $U_{i+1}(d_{i+1})$ is convex in
$\ve{z}_{\ve{x}_{0:i+1}}$. We have to prove that
$U_{i}(d_{i})$ is convex in $\ve{z}_{\ve{x}_{0:i}}$.

From (\ref{eq:4r4}), the expectation under
the normal variable $Z_{x_{i+1}}$ with posterior mean $\mu_{Z_{x_{i+1}}|d_i}$ and variance $\sigma^2_{Z_{x_{i+1}}|d_i}$ can be expressed in terms of
the standard normal variable $Z = (Z_{x_{i+1}}-\mu_{Z_{x_{i+1}}|d_i})/\sigma^2_{Z_{x_{i+1}}|d_i}$:
$$
\begin{array}{l}
\displaystyle\int f(Z_{x_{i+1}}=z_{x_{i+1}}| d_i) \ U_{i+1}(d_{i}, x_{i+1}, z_{x_{i+1}}) \ \mbox{d}z_{x_{i+1}} =\\
\displaystyle\int f(z) \ U_{i+1}(d_{i}, x_{i+1}, \mu_{Z_{x_{i+1}}|d_i}+ \sigma^2_{Z_{x_{i+1}}|d_i}z) \ \mbox{d}z \ .
\end{array}
$$

Since $d_{i}$ and $\mu_{Z_{x_{i+1}}|d_i}+ \sigma^2_{Z_{x_{i+1}}|d_i}z$ are affine in $\ve{z}_{\ve{x}_{0:i}}$ and $U_{i+1}(d_{i+1})$ is convex in
$\ve{z}_{\ve{x}_{0:i+1}}$ by assumption, $U_{i+1}(d_{i}, x_{i+1}, \mu_{Z_{x_{i+1}}|d_i}+ \sigma^2_{Z_{x_{i+1}}|d_i}z)$ is convex in $\ve{z}_{\ve{x}_{0:i}}$ because vector composition operation preserves convexity\footnote{Note that $U_{i+1}(d_{i+1})$ does not have to be non-decreasing in each argument because $d_{i}$ and $\mu_{Z_{x_{i+1}}|d_i}+ \sigma^2_{Z_{x_{i+1}}|d_i}z$ are affine in $\ve{z}_{\ve{x}_{0:i}}$.} ((Boyd and Vandenberghe 2004), pp. 86). Since $U_{i+1}(d_{i}, x_{i+1}, \mu_{Z_{x_{i+1}}|d_i}+ \sigma^2_{Z_{x_{i+1}}|d_i}z)$ is convex in $\ve{z}_{\ve{x}_{0:i}}$ for each $z$, $\displaystyle\int f(z) \ U_{i+1}(d_{i}, x_{i+1}, \mu_{Z_{x_{i+1}}|d_i}+ \sigma^2_{Z_{x_{i+1}}|d_i}z) \ \mbox{d}z$ is convex in $\ve{z}_{\ve{x}_{0:i}}$ because integration preserves convexity ((Boyd and Vandenberghe 2004), pp. 79).
So, $\displaystyle\int f(z_{x_{i+1}}| d_i) \ U_{i+1}(d_{i}, x_{i+1}, z_{x_{i+1}}) \ \mbox{d}z_{x_{i+1}}$ is convex in $\ve{z}_{\ve{x}_{0:i}}$.
From above, $\mathbb{H}[Y_{x_{i+1}}| d_i]$ is convex in $\ve{z}_{\ve{x}_{0:i}}$.
Then, the pointwise maximum of $\displaystyle \mathbb{H}[Y_{x_{i+1}}| d_i] + \int f(z_{x_{i+1}}| d_i) \ U_{i+1}(d_{i}, x_{i+1}, z_{x_{i+1}}) \ \mbox{d}z_{x_{i+1}}$ is convex in $\ve{z}_{\ve{x}_{0:i}}$. Therefore, $U_{i}(d_{i})$ is convex in $\ve{z}_{\ve{x}_{0:i}}$. The inductive case is true.
\subsection{Theorem~\ref{thm:9}}
\label{sect:app5}
\emph{Proof by induction} on $i$ that $\underline{U}^{\nu}_{i}(d_i) \leq \underline{U}^{\nu + 1}_{i}(d_i) \leq U_{i}(d_i)$ for $i = t,\ldots,0$.\\

\noindent
\emph{Base case} ($i = t$): $\underline{U}^{\nu}_{t}(d_t) = \underline{U}^{\nu + 1}_{t}(d_t) = U_{t}(d_t) = \displaystyle\max_{\ve{a}_{t} \in \set{A}'(\ve{x}_t)}
\mathbb{H}[Y_{x_{t+1}}| d_t]$. Hence, the base case is true.\\

\noindent
\emph{Inductive case}: Suppose that 
$\underline{U}^{\nu}_{i+1}(d_{i+1}) \leq \underline{U}^{\nu + 1}_{i+1}(d_{i+1}) \leq U_{i+1}(d_{i+1})$
is true. We have to prove that 
$\underline{U}^{\nu}_{i}(d_i) \leq \underline{U}^{\nu + 1}_{i}(d_i) \leq U_{i}(d_i)$ is true.

We will first show that $\underline{U}^{\nu + 1}_{i}(d_i) \leq U_{i}(d_i)$.
$$
%\hspace{-2mm}
\begin{array}{l}
\underline{U}^{\nu + 1}_{i}(d_i)\\
= \displaystyle\max_{\ve{a}_{i} \in \set{A}'(\ve{x}_i)}
\mathbb{H}[Y_{x_{i+1}}| d_i] +
\sum^{\nu +1}_{j=1} \underline{p}^{[j]}_{x_{i+1}} \underline{U}^{\nu + 1}_{i+1}(d_i, x_{i+1}, \underline{z}^{[j]}_{x_{i+1}}) \\
\leq \displaystyle\max_{\ve{a}_{i} \in \set{A}'(\ve{x}_i)}
\mathbb{H}[Y_{x_{i+1}}| d_i] +
\sum^{\nu + 1}_{j=1}  \underline{p}^{[j]}_{x_{i+1}} {U}_{i+1}(d_i, x_{i+1}, \underline{z}^{[j]}_{x_{i+1}})\\
\leq \displaystyle\max_{\ve{a}_{i} \in \set{A}'(\ve{x}_i)} 
\mathbb{H}[Y_{x_{i+1}}| d_i] + \mathbb{E}[U_{i+1}(d_i, x_{i+1}, Z_{x_{i+1}})| d_i]\\
= U_{i}(d_{i}) \ .
\end{array}
$$ 
The first inequality follows from assumption (i.e.,
$\underline{U}^{\nu + 1}_{i+1}(d_i, x_{i+1}, \underline{z}^{[j]}_{x_{i+1}}) \leq U_{i+1}(d_i, x_{i+1}, \underline{z}^{[j]}_{x_{i+1}})$). 
The second inequality follows from Lemma~\ref{lem:3} that $U_{i+1}(d_i, x_{i+1}, z_{x_{i+1}})$ is convex in $z_{x_{i+1}}$ for $\ell$GP, and the generalized Jensen bound (\ref{eq:5v}).

We will now prove that $\underline{U}^{\nu}_{i}(d_i) \leq \underline{U}^{\nu + 1}_{i}(d_i)$.
$$
\begin{array}{l}
\underline{U}^{\nu}_{i}(d_i)\\ 
= \displaystyle\max_{\ve{a}_{i} \in \set{A}'(\ve{x}_i)}
\mathbb{H}[Y_{x_{i+1}}| d_i] +
\sum^{\nu}_{j=1} \underline{p}^{[j]}_{x_{i+1}}
\underline{U}^{\nu}_{i+1}(d_i, x_{i+1},  \underline{z}^{[j]}_{x_{i+1}}) \\
\leq \displaystyle\max_{\ve{a}_{i} \in \set{A}'(\ve{x}_i)}
\mathbb{H}[Y_{x_{i+1}}| d_i] +
\sum^{\nu}_{j=1} \underline{p}^{[j]}_{x_{i+1}}
\underline{U}^{\nu + 1}_{i+1}(d_i, x_{i+1},  \underline{z}^{[j]}_{x_{i+1}}) \\
\leq \displaystyle\max_{\ve{a}_{i} \in \set{A}'(\ve{x}_i)} 
\mathbb{H}[Y_{x_{i+1}}| d_i] + \sum^{\nu +1}_{\ell=1} \underline{p}^{[\ell]}_{x_{i+1}} \underline{U}^{\nu + 1}_{i+1}(d_i, x_{i+1},  \underline{z}^{[\ell]}_{x_{i+1}})\\
= \underline{U}^{\nu + 1}_{i}(d_i) \ .
\end{array}
$$
The first inequality follows from assumption (i.e.,
$\underline{U}^{\nu}_{i+1}(d_i, x_{i+1}, \underline{z}^{[j]}_{x_{i+1}}) \leq \underline{U}^{\nu + 1}_{i+1}(d_i, x_{i+1}, \underline{z}^{[j]}_{x_{i+1}})$).
We need the result that $\underline{U}^{\nu+1}_i(d_i)$ is convex in
$\ve{z}_{\ve{x}_{0:i}}$ for $i = 0, \ldots, t$ for the second inequality to hold. The proof\footnote{The approximate problems $\underline{i}$MASP($\frac{1}{k}$) and $\overline{i}$MASP($\frac{1}{k}$) differ from $i$MASP($\frac{1}{k}$) (\ref{eq:4r4}) by the non-negative weighted sum (instead of the expectation), which also preserves convexity.} is similar to that of Lemma~\ref{lem:3}.
Consequently, since $\underline{U}^{\nu + 1}_{i+1}(d_i, x_{i+1}, z_{x_{i+1}})$ is convex in $z_{x_{i+1}}$ and $\set{Z}^{\nu+1}_{x_{i+1}}$ is obtained by splitting one of the intervals in $\set{Z}^{\nu}_{x_{i+1}}$, the second inequality results.
The inductive case is thus true.

The proof of $U_{i}(d_i) \leq \overline{U}^{\nu + 1}_{i}(d_i) \leq \overline{U}^{\nu}_{i}(d_i)$ for $i = t, \ldots, 0$ is similar to the above except that the inequalities are reversed.

\subsection{References for Proofs}
{\small
Boyd, S., and Vandenberghe L. 2004. \emph{Convex Optimization}. Cambridge Univ. Press.\vspace{1mm}

\noindent
Cover T., and Thomas J. 1991. \emph{Elements of Information Theory}. John Wiley $\&$ Sons.
}
\end{document}